\documentclass[a4paper,fleqn]{cas-sc}

\usepackage[authoryear,longnamesfirst]{natbib}
\usepackage{float}
\usepackage[section]{placeins}

\usepackage{caption}
\usepackage{subcaption}

\usepackage{todonotes}

\def\tsc#1{\csdef{#1}{\textsc{\lowercase{#1}}\xspace}}
\tsc{WGM}
\tsc{QE}
\tsc{EP}
\tsc{PMS}
\tsc{BEC}
\tsc{DE}

\begin{document}
\let\WriteBookmarks\relax

\shorttitle{How Valid Are Thermal Comfort Cohort Models}

\shortauthors{Hirsch et~al.}

\title [mode = title]{Replication of Thermal Comfort Cohort Models: No Support for Improved Thermal Preference Prediction}                      


%

%
\author[1,2]{Thomas Hirsch}

\cormark[1]





\affiliation[1]{organization={TU Wien},
    city={Vienna},
    country={Austria}}

\author[1]{Katja Ceplak}

\author[1]{Veronika Windisch}

\author[1]{Samina \text{Kadkhoda Masoumali}}


\affiliation[2]{organization={Complexity Science Hub},
    city={Vienna},
    country={Austria}}

\author[3,4]{Matias Quintana}

\affiliation[3]{organization={Singapore Management University}, 
    city={Singapore},
    country={Singapore}}
    
\affiliation[4]{organization={Singapore-ETH Centre},
    city={Singapore},
    country={Singapore}}

\author[3]{Clayton Miller}

\author[5]{Stefano Schiavon}

\affiliation[5]{organization={University of California, Berkeley},
    city={Berkeley},
    country={USA}}

\author[1,2]{Gerald Schweiger}


\cortext[cor1]{Corresponding author}



\begin{abstract}
Cohort comfort models (CCMs) aim to overcome the extensive individual data requirements of personal comfort models (PCMs) by grouping occupants based on demographic and psychometric characteristics (cold start) or similar thermal preferences (warm start).
We evaluated the validity of the reported CCM performance gains \cite{quintanaCohortComfortModels2023} through (i) a methodological assessment of the original study, (ii) a replication using one of the two original datasets (DORN: 20 participants, 6 months) under a revised experimental protocol, and (iii) a replication using an independently collected dataset (PersonAI: 72 participants, 4 weeks).
Classification performance was evaluated using F1-micro, as in the original study, and F1-macro to account for class imbalance.
Under F1-micro, a trivial baseline that always outputs the majority class performed on par with cohort models.
Under F1-macro, all cohort models achieved lower median scores than a simple baseline that makes random predictions based on the estimated distribution of thermal preferences.
In our methodological assessment we identified the following key issues: Limited data per participant and per minority class, reliance on a metric insensitive to class imbalance, missing trivial baselines, inadequate cross validation strategy, and multiple occurrences of information leakage.
Our findings challenge the original study’s conclusion that CCMs improve thermal preference prediction. 
CCMs did not outperform simple class-distribution baselines.
Future studies should improve experimental designs, including controlled thermal conditions, to obtain sufficient observations across thermal preference classes.
\end{abstract}




\maketitle

\section{Introduction}

Buildings face a dual and often conflicting challenge. 
They play a central role in global efforts to reduce greenhouse gas emissions, accounting for approximately 40\% of worldwide energy demand \cite{schweiger2020active}.
At the same time, they must provide indoor conditions that support health, comfort, and performance, as people spend the vast majority of their lives indoors \cite{klepeis2001national}.
Over the past decades, researchers and standardization bodies have developed methods to predict indoor environmental quality and occupant comfort that are built on aggregated representations of how a typical occupant perceives the thermal environment \cite{donkers2025personal}. 
These aggregated representations are inherently limited because they estimate group averages and therefore often predict poorly for individuals. 
As a result, designing and operating buildings according to such models can fail to improve real occupant satisfaction and may instead encourage conservative HVAC sizing and scheduling, increasing energy use without reliably reducing discomfort \cite{ donkers2025personal}.
To address the limitations of aggregate models, data-driven Personal Comfort Models (PCMs) have been proposed.
PCMs are a new approach to thermal comfort modeling that predicts individuals' thermal comfort responses, instead of the average response of a large population and it can be applied to any HVAC system \cite{kimPersonalComfortModels2018a}.
The increasing availability of off-the-shelf wearable devices has strengthened this approach by enabling the collection of physiological measurements, such as skin temperature and heart rate, which have been shown to improve prediction accuracy. 
Yet collecting longitudinal field data remains costly and burdensome, as it requires repeated Right-Here-Right-Now (RHRN) surveys during daily activities and high-resolution environmental monitoring that is difficult to sustain at scale \cite{tartariniPersonalComfortModels2022}.
To address the scalability limitations of personalized models, Cohort Comfort Models (CCMs) have been proposed, exploiting similarities between occupants to estimate the thermal preferences of new individuals with little or no historical data \cite{quintanaCohortComfortModels2023}.
By estimating preferences based on similarities between occupants, CCMs combine the personalization of individual models with the data efficiency of aggregate approaches.

If CCMs are to inform real design and operational decisions at scale, their empirical claims must be sufficiently robust. 
At present, no independent replication has evaluated these claims, and the statistical methods and machine learning procedures used in the original study have not yet been independently re-examined for reproducibility, robustness, or potential sources of bias.

\subsection{Main Contribution}

To evaluate the credibility of the CCMs, we pursue a multi-level validation: 
\begin{itemize}

\item Methodological assessment. We first reviewed the original experimental setup, including the data splitting strategy, model training procedure, and evaluation protocol. Based on this assessment, we revised the implementation to conform to established Machine Learning (ML) best practices and to eliminate methodological shortcomings that could compromise the validity of the reported findings. The resulting protocol, which preserves the original research objective while correcting the identified implementation shortcomings, is hereafter referred to as the \emph{revised experimental setup}.

\item Replication using the original data.
We examined whether the original conclusions hold when the study is repeated using the revised experimental setup and one of the original datasets. We also requested access to the second dataset used in the original study and contacted the original authors; however, this dataset was unavailable

\item Replication using independently collected data, generating the PersonAI dataset.
We examined whether the original conclusions generalize to an independent dataset collected through a field experiment involving 80 participants, using the revised experimental setup.
\end{itemize}

The two replication studies preserve the original research hypothesis and experimental design while introducing deliberate changes to the implementation. 
Under the framework of Desai et al.~\cite{desai2025reproducibility}, they are classified as \emph{direct replications}, whereas under that of Gundersen~\cite{gundersen2021fundamental}, they are classified as \emph{interpretation reproducibility} studies.
\section{Background}

Academic success depends heavily on publication records, which influence funding \cite{bol2018matthew}, promotion \cite{horta2016impact}, and tenure \cite{tripodi2025tenure}. However, scientific publishing favors novel, positive, and unambiguous findings over negative results and replication studies \cite{nosek2012scientific}. Together, these factors create structural incentives for researchers to prioritize novel and positive findings over replication and rigorous methodological scrutiny.
This current practice is defensible only under the implicit assumption that the large majority of published discoveries are correct \cite{ioannidis2018replication}. 
If science were to produce numerous major discoveries, most of which are true and accurate, allocating resources to replication would indeed appear to be of low priority. 
However, empirical evidence across diverse scientific fields suggests that this narrative is untenable \cite{camerer2016evaluating, ioannidis2005most,errington2021investigating}. 
Science is a difficult endeavor, and numerous biases affect different study designs and disciplines \cite{fanelli2017meta}; even with the best intentions, findings can easily be false, exaggerated, or inflated. 
In scientific systems where genuine discoveries are not abundant and the proportion of false findings is appreciable, replication may have greater scientific value than an additional original discovery claim. 
Moreover, if false findings carry substantial consequences (such as wasted resources, misdirected research efforts, or harm to end users), original discovery work may even have negative value \cite{ioannidis2018replication}.
Beyond the structural disincentives that discourage replication efforts, a second and independent challenge concerns the practical feasibility of reproducibility itself. 
Even when researchers attempt to verify prior findings, reproduction may fail because methods are not described in sufficient detail, analyses are incompletely or selectively reported, or the underlying data are unavailable \cite{munafo2017manifesto}.
In AI research, these general problems are compounded by additional barriers, including under-specified model architectures and training procedures, missing information about data preprocessing, selective reporting of results, and the absence of code and data \cite{gundersen2021fundamental, semmelrock2025reproducibility}. 
Some contemporary AI models also require substantial computational infrastructure for training and evaluation; although reproducible in principle, their cost and scale restrict independent verification and thereby limit reproducibility in practice.

As in many other applied engineering disciplines, research on buildings has closely followed the recent wave of enthusiasm for AI \cite{li2025machine}.
AI has been applied to system control \cite{yang2020model}, fault detection \cite{mukhtar2024one}, and energy prediction \cite{schranz2021energy}, as well as to the modelling of IEQ, including personalised thermal preference modelling \cite{jayathissa2020humans}.
While replication studies in engineering remain relatively rare, systematic investigations into the reproducibility of machine learning research have begun to emerge \cite{hong2026good, mukhtar2025reproducibility}.

\subsection{From Personal to Cohort Comfort Models}
The Predicted Mean Vote (PMV) model, introduced by Fanger~\cite{fanger1970}, has long served as the standard for thermal comfort assessment in buildings. 
Designed to predict the thermal sensation of a group of persons, the PMV acknowledges individual variation but sets it aside. It is also used to predict the comfort of a single person, for example someone occupying a private office.  
A related population-level approach, the Adaptive model~\cite{deDearBrager1998}, similarly averages individual responses by linking comfort to prevailing outdoor temperatures. 
Both models have proven useful at the aggregate level, yet studies using the ASHRAE Global Thermal Comfort Database II~\cite{ashraeDatabaseII2018} reveal that the PMV achieves only 33\% accuracy when predicting the thermal sensation of a specific individual~\cite{cheungPMVAccuracy2019}; an unsatisfying result for occupant-centric applications.
 The proliferation of IoT sensors and wearable devices opened the door to a more granular alternative: PCMs ~\cite{kimPersonalComfortModels2018a}. 
By pairing direct occupant feedback with environmental and/or physiological and/or behavioural data collected longitudinally in the field, PCMs can be tailored to an individual's specific preferences and have been shown to substantially outperform the PMV at the personal level~\cite{kimPersonalComfortModels2018}.
Despite their promise, PCMs carry a significant practical burden: collecting the 60--200 labeled data points per occupant required for stable rediction~\cite{kimPersonalComfortModels2018, liPersonalized2017, daumPersonalized2011, liuPersonalThermalComfort2019} is logistically demanding, disruptive to daily activities, and cost-prohibitive at scale, particularly in flexible workspaces where occupancy turnover is high.
 CCMs ~\cite{quintanaCohortComfortModels2023} are introduced as a middle path between these two extremes.
Rather than averaging across an entire population or waiting to accumulate sufficient individual data, CCMs identify groups of occupants who share underlying thermal preference patterns and leverage their pooled historical data to generate predictions for a new arrival. 
By drawing on onboarding surveys, physiological and environmental sensor
measurements, and as few as a single labeled comfort response, CCMs seek to capture the personalization benefits of PCMs without the associated data collection burden.

\subsection{Cohort Comfort Models}
Building on the personal comfort modeling framework established by Kim et al.~\cite{kimPersonalComfortModels2018, kimPersonalComfortModels2018a}, Quintana et al.~\cite{quintanaCohortComfortModels2023} addressed a fundamental limitation of PCMs: the requirement to collect a large number of labeled data points from every individual occupant before a useful model can be trained.
Where Kim et al.\ demonstrated that roughly 60 survey responses per occupant are needed for stable prediction, Quintana et al.\ asked whether historical data from \emph{similar} occupants could substitute for (or significantly reduce) that individual data collection burden.
To this end, they introduced CCMs, a framework that clusters occupants into groups based on shared thermal preference characteristics and uses the resulting group model to predict the comfort of a new arrival with little or no personal history.
The framework was validated on two longitudinal datasets totalling 55 participants and more than 6000 RHRN thermal preference surveys: the DORN dataset (20 occupants, Singapore, 6 months) and the SMC dataset (37 occupants, Redwood City CA, 12 weeks).
The cohort creation followed two strategies.
In the \emph{cold start} scenario (directly analogous to the onboarding challenge Kim et al.\ identified) cohorts were formed using only one-time survey responses (demographic data, the Satisfaction With Life Scale, the Highly Sensitive Person Scale, and the Big Five Personality inventory), requiring no longitudinal comfort data from the new occupant whatsoever.
In the \emph{warm start} scenario, cohorts were constructed from cross-model performance evaluation and response distribution similarity between occupants' PCMs, and a new occupant was assigned to a cohort using as few as a single labeled RHRN data point.
Random Forest classifiers, the same algorithm identified as top-performing by Kim et al. ~\cite{kimPersonalComfortModels2018}, were used throughout, and model performance was evaluated with the F1-micro score.
They found that warm start approaches consistently outperformed both the general-purpose aggregate model and the cold start approaches, achieving median F1-micro scores of 0.61--0.72 across the two datasets.
At the occupant level, they found that roughly half of participants in the DORN dataset and one third in the SMC dataset experienced an average improvement of 8\% and 5\% respectively over the general-purpose baseline, with gains as large as 36\% and 46\% for the most responsive individuals.
Among cold start approaches, they found that cohorts based on the Satisfaction With Life Scale and the Agreeableness personality trait provided modest but consistent improvements over the general-purpose model, suggesting that even purely demographic and psychometric information can meaningfully stratify occupant thermal preferences.
They conclude that CCMs can match or exceed the predictive accuracy of general-purpose data-driven models while requiring as little as a single labeled observation from the new occupant, substantially reducing the data collection burden that Kim et al. ~\cite{kimPersonalComfortModels2018} had identified as the primary practical obstacle to personalized thermal comfort modeling at scale.

\section{Method}
\subsection{Study Design and Data Collection}
\begin{figure}
    \centering
    \includegraphics[width=\linewidth]{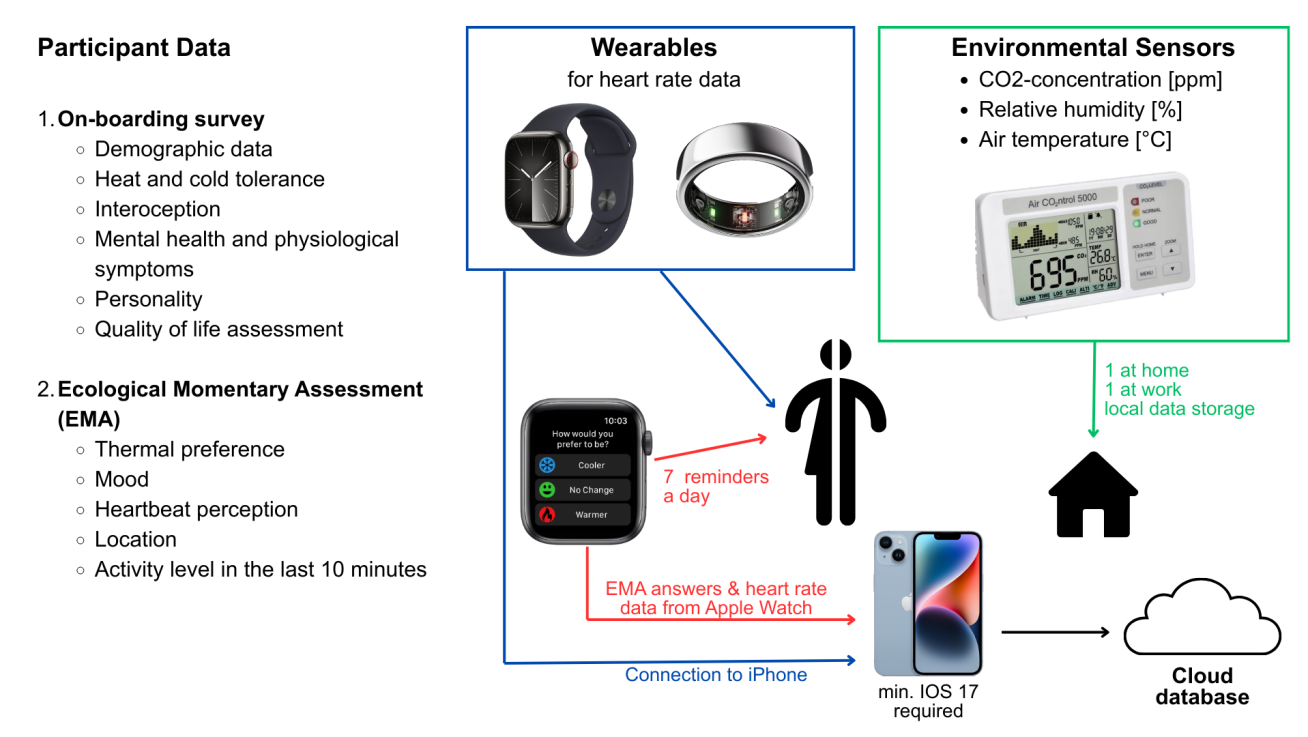}
    \caption{Overview of the study methodology.}
    \label{fig:methodology_overview}
\end{figure}

Over a four-week period, we collected subjective, physiological, and environmental data to investigate thermal comfort under real-life conditions. Different wearable devices and sensors were used for this purpose (Figure~\ref{fig:methodology_overview}). 
The data were then used to train the CCMs, which established separate groups based on different thermal preferences.
This study was approved by the ethical committee of the Graz University of Technology, Austria (GZ: EK-35/2024). 
As a participation compensation, the participants kept the Apple Watches used in the study.

\subsubsection{Participants}
A total of 83 individuals volunteered to participate in the study. Participants were excluded from the final data set if they did not complete the four-week data collection, if their environmental or physiological data were missing, or if the timestamps from different devices did not match. 
Therefore, the total sample consisted of 72 participants, of whom 40 were women (age: M = 32.3, SD = 9.0, range = 18 - 64). The sample was highly educated: nine participants (12.5\%) held a PhD, 28 (38.9\%) a master's degree, and 20 (27.8\%) a bachelor's degree, while 14 (19.4\%) had completed upper secondary education and one (1.4\%) compulsory schooling only.
Based on the Body-Mass-Index and self-report, participants were healthy (BMI; M = 23.4, SD = 3.4 ) and the majority reported participating in regular exercise (92\%). Living arrangements were skewed towards living in an apartment (75\%), followed by a single-family home (19\%) and, lastly, a townhouse (6\%)
The sample was recruited through social media and word of mouth. The inclusion criteria for participation were 18 to 65 years of age, working in an office setting and owning an iPhone with an IOS 17 operating system or later, which supported an Apple Watch Series 9. The exclusion criteria consisted of having a chronic condition and taking medication that might interfere with perception of air quality or thermoregulation. After potential participants disclosed their chronic conditions and medications in the screening survey, the evaluation was done on a case-by-case basis.

\subsubsection{Wearable sensors and devices}
Each participant was equipped with an Apple Watch, an Oura Ring, and two environmental sensors.

\textbf{Apple Watch}. Apple Watches Series 9 were used in the study. 
The watches were coupled with the participants' own iPhones. 
The Cozie app \cite{Tartarinietal2023} was used to present RHRN surveys directly on the Apple Watch screen and to obtain heart rate (HR) data collected by the Apple Watch. Participants were instructed to wear their watch during the scheduled RHRN surveys, from 08:00 to 20:00 every day.

\textbf{Oura Ring}. In addition to smart watches, Oura Rings Gen4 were also used in the study. 
In order for the smart ring to collect data, the Oura app needed to be installed on the participants' own phone. 
Participants were instructed to wear the ring as much as possible, however, they were free to remove it at night if discomfort occurred. 
No restrictions were made on the visibility and accessibility of the app features. HR data from the Oura Ring were used as the primary physiological input for the comfort models, as they were available at a higher temporal resolution and more consistently throughout the day, including around RHRN prompts. The smart ring measures daytime HR for one minute every five minutes \cite{OuraHealth2026} whereas the Apple Watch records background HR at irregular intervals depending on the person's activity \cite{AppleInc2026}. Oura-derived HR has also shown high correspondence with a research-grade reference for resting HR under real-life conditions \cite{Henriksenetal2022}.

\textbf{AirCo2ntrol 5000}. Each participant was equipped with two indoor air quality sensors. The AirCo2ntrol 5000 device \cite{TFADostmannnd}, was used to measure temperature, relative humidity, and CO$_2$-concentration in 5 second intervals. Participants were instructed to place one sensor on their work desk and one in their home, specifically in the room where they spent most of their leisure time. They were instructed to not change the location of the sensor for the duration of the study and to not place them in direct sunlight. The sensor did not need manual calibration. It was chosen because of its user-friendliness that allowed the participants to configure the sensor on their own. Due to the participants placing the sensor at their place of work, we opted out of wireless data-transfer for data protection reasons.  Data were stored on a SD-card. 

\subsubsection{Questionnaires}
Self-report data were collected via two sources: RHRN and an onboarding surveys. 
The RHRN surveys provided subjective data on the experience of current thermal conditions and contextual information. 
The onboarding survey provided demographic data, as well as heat and cold tolerance, interoception, mental health and physiological symptoms, personality, and satisfaction with life assessments.

\textbf{Right-Here-Right-Now}. 
The RHRN survey consisted of five questions that were presented seven times per day, every day between 08:00 and 20:00 in a 105-min interval. This totals to 196 surveys over the span of four weeks. Participants answered the surveys on their Apple Watch screen through the Cozie app \cite{Tartarinietal2023}, allowing them to answer quickly, without much effort or disruption. 

The RHRN survey consisted of:
\begin{enumerate}
    \item How would you prefer to be?
       
    Answers: cooler, no changes, warmer
    \item How do you feel right now?
       
    A: good, neutral, bad
    \item In this moment, I can feel my heart beat: 
       
    A: very strongly, strongly, medium, weak, not at all
    \item Where are you?
      
    A: at home, at work, on the go
    \item What did you do in the last 10 min?
      
    A: physical activity, desk work, relaxation, other
\end{enumerate}

The questions were used to assess momentary thermal preference (1), current mood (2), body awareness, specifically heartbeat perception (3, \cite{Walentynowiczetal2023}), location and proximity to an environmental sensor (4) and activity level in the last 10 min (5). The questions were always presented in the same order. If an inquiry was missed, an RHRN could be manually triggered anytime by opening the Cozie app on the Apple Watch. Participants were instructed to complete a minimum of 147 RHRN surveys (75\%) to meet the participation compensation requirement.
A question about clothing was not included, as previous work found that adding self-reported clothing information did not consistently improve the predictive performance of models \cite{tartariniPersonalComfortModels2022}.

\textbf{Heat and cold tolerance}. 
Participants' self-reported heat and cold tolerance was accessed through a six item questionnaire. 
The self-constructed questionnaire consisted of three items dedicated to each temperature tolerance. 
Examples were "I find high temperatures very unpleasant" for heat tolerance and "I tend to freeze faster than others" for cold tolerance. 
Participants answered on a five-point Likert scale ranging from "1 - not applicable at all" to "5 - fully applicable". The internal consistency was acceptable for the cold, but questionable for the heat scale (cold: McDonald's omega = .76, heat: McDonald's omega =.64).

\textbf{Interoception}. 
The Multidimensional Assessment of Interoceptive Awareness (MAIA-2 \cite{Mehlingetal2018}) was used to assess self-reported interoceptive awareness by trait. 
The questionnaire consisted of eight scales: Noticing, Not-Distracting, Not-Worrying, Attention Regulation, Emotion Awareness, Self-Regulation, Body Listening, and Trust. The items were answered on a six-point Likert scale, ranging from never (0) to always (5). An example of the Not Worrying scale is: “I can notice an unpleasant body sensation without worrying about it”. The internal consistency was satisfactory (McDonalds's omega = .80 - .93).

\textbf{Symptoms}. 
The Brief Symptom Inventory (BSI \cite{Derogatis1975} \cite{Franke2000}) assessed mental health and psychological symptoms through 53 items that were merged into nine scales: Somatization, Obsession-Compulsion, Interpersonal Sensitivity, Depression, Anxiety, Hostility, Phobic anxiety, Paranoid ideation, and Psychoticism. Items were answered on a five-point Likert scale, ranging from not at all (0) to extremely (4). The scales showed adequate internal consistency (McDonald's omega = .77 - .88).

\textbf{Personality}. 
The Big Five Personality Inventory (BFI-2-S \cite{Sotojohn2017} \cite{Rammstedtetal2020}) examined five personality traits Extraversion, Agreeableness, Conscientiousness, Negative Emotionality and Open-Mindedness with 30 items, six per dimension. Items were answered on a five-point Likert scale, ranging from strongly disagree (1) to strongly agree (5). “I am someone who has difficulty getting started on tasks” is one example of an item. Internal consistency was acceptable (McDonald's omega = .74 - .89).

\textbf{Satisfaction with life}. 
Satisfaction with Life Scale (SWLS \cite{Dieneretal1985} \cite{JankeGloecknerRist2012}) consisted of 5 items, which were answered on a seven-point Likert scale, ranging from strongly disagree (1) to strongly agree (7). 
The instrument assesses cognitive-evaluative and affective components of life satisfaction. 
An example of an item is: “If I could live my life over, I would change almost nothing”. Internal consistency was sufficient (McDonald's omega = .89).

\subsubsection{Procedure}
The data collection procedure took place in four separate steps. 
People who were interested in participating in the study first completed the screening survey in which the inclusion and exclusion criteria were checked. Suitable participants were invited to an onboarding event where consent forms were signed, wearables and sensors were distributed and set-up, and tasks were explained. 
Afterwards, the four-week data collection started in which the participants filled out seven RHRN surveys per day. Participants were also asked to complete the onboarding survey shortly after the start of the study.

\subsection{Machine Learning: Revised Experimental Setup}
As this work is a replication study, we follow the machine learning methodology proposed by Quintana et al. \cite{quintanaCohortComfortModels2023}. 
The original experimental design, model architecture, and cohorting framework are therefore not described again here. 
Instead, this section focuses exclusively on the methodological and implementation changes introduced in our revised experimental setup.

The revisions were motivated by three objectives:
\begin{itemize}
    \item correcting methodological shortcomings,
    \item more robust assessment of classification performance in regards to class imbalances,
    \item fixing implementation errors (e.g., information leakage and dataset loading issues), and
    \item reducing unnecessary computational overhead.
\end{itemize}

\textbf{Train/Test splitting}.
Instead of 100 repetitions of randomized 80/20 train test splits we utilized a leave-one-out cross-validation process.
The reason for this is two fold:
First, the original approach is statistically weak and such obtained results for such small datasets are hard to interpret.
Second, the 100 repetitions of random splits of datasets containing less than 50 persons does not increase confidence in the results, especially given the high variance of the obtained results, and does not warrant the computational overhead.
To summarize, results obtained from a leave-one-out cross-validation process are easier to interpret while in this special case causing less computational overhead.

\textbf{Performance metrics}.
The original publication evaluates classification performance using micro averaged F1, which is sensitive to class imbalance.
On imbalanced datasets such as DORN and PersonAI, it can therefore yield inflated and potentially misleading results.
We therefore als include macro averaged F1 to provide a more robust evaluation.

\textbf{Hyperparameter tuning}.
We omitted hyperparameter optimization and evaluated all models using the default  parameters. In the original experimental setup, hyperparameters were optimized independently for each of the 100 randomized train–test splits, resulting in a different model specification for each repetition. This simplification also substantially reduced the computational cost of the experiments.

\textbf{Dataset handling}.
The DORN dataset in its provided form\footnote{\url{https://github.com/FedericoTartarini/ dorn-longitudinal-tc-study}} could not be read with the with the provided CCM code\footnote{\url{https://github.com/buds-lab/ccm.}}.
Adaptions to the dataset loading code were required in order to be able to load the dataset and run the experiments.
As the original paper include a dataset version in the form of a repository release, tag, or commit hash, it was not possible to identify the exact software state used to produce the original results.

\textbf{Feature selection logic}.
In the original codebase, model inputs were defined by excluding features rather than explicitly specifying the features to include. This approach is prone to errors because newly introduced variables can unintentionally become part of the model input. During our review, we identified a bug that caused information leakage by inadvertently including a feature that was not intended for the corresponding experiment. We therefore replaced the exclusion-based logic with explicit feature selection.

\subsection{Baselines}
We introduce a number of baselines to contextualize and compare the measured performance scores.
The \emph{Majority} classifier does not look at any input features during training, but instead simply determine the majority class found in the training set. On inference, this class is predicted always. For example, the class distribution in the training set is 70\% "no change", 20\% "prefer cooler" and 10\% "prefer warmer", the classifier will always return "no change".
The \emph{PCM Majority} is such a classifier as described above and was "trained" and evaluated alongside the PCM models.
The \emph{Cohort Majority} is such a classifier as described above and was "trained" and evaluated alongside the cohort models.
The \emph{ApproxDistribution} classifier does not look at any input features during training, but instead estimates the target distribution by sampling $10$ points from the training set. On inference, a random class value is produced adhering to the learned distribution. For example, 10 points are randomly sampled from the trainig set, there are 7 "no change", 2 "prefer cooler" and 1 "prefer warmer", the classifier will return a random class, with the probability of 0.7 for for "no change", 0.2 for "prefer cooler" and 0.1 for "prefer warmer.
The \emph{Random F1 score for a balanced 3-class problem}, being the F1 score produced by a uniformly random choices of classes (uniform distribution) used as ground truth and its predictor, being 0.333 F1 macro/micro.
\section{Results}

\subsection{Correlation Analysis}
\label{sec:correlation_analysis}
\begin{figure}[htbp]
    \centering
    \includegraphics[width=0.5\textwidth]{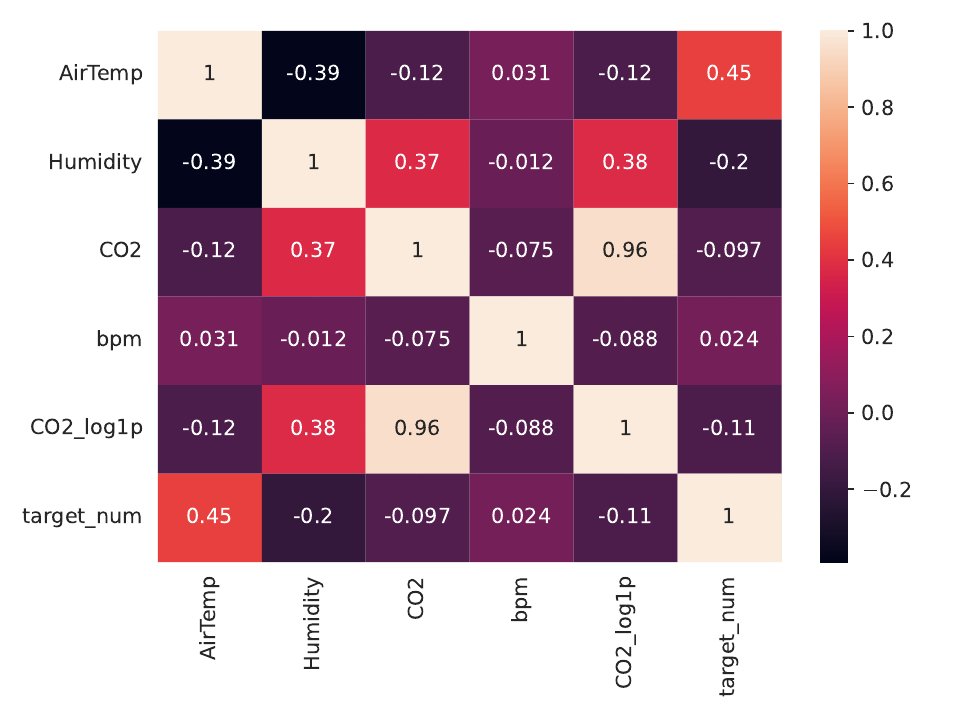}
    \caption{Pearson correlation matrix between numeric environmental/biological features and the thermal comfort target. }
    \label{fig:correlation_pearson_numeric_target}
\end{figure}

Correlation analysis on the PersonAI dataset (see Figure~\ref{fig:correlation_pearson_numeric_target}) showed that temperature had the strongest linear association with thermal preference (Pearson's \(r = 0.45\)). 
The remaining environmental and physiological variables were more weakly associated with the outcome: relative humidity (\(r = -0.20\)), CO$_2$ concentration (\(r = -0.097\)), its log-transformed counterpart (\(r = -0.11\)), and heart rate (\(r = 0.024\)).

Correlation analysis of the onboarding variables (see Table~\ref{table:correlation_pearson_onboarding_target}) showed generally weak linear associations with thermal preference. 
The strongest positive associations were observed for Openness (BFI2S\_Offenheit; Pearson's \(r = 0.23\)), its short-form counterpart (BFI2XS\_Offenheit; \(r = 0.21\)), and related openness facets, including aesthetic sensitivity (\(r = 0.18\)), intellectual curiosity (\(r = 0.18\)), and creative imagination (\(r = 0.16\)). 
The strongest negative associations were observed for the anxiety facet (BFI2S\_Ängstlichkeit; \(r = -0.20\)), MAIA Trust (\(r = -0.17\)), and the short-form Negative Emotionality scale (\(r = -0.17\)). 
The Satisfaction With Life Scale (SWLS), one of the onboarding variables used for cold-start cohort assignment in the original CCM framework, showed only a weak linear association with thermal preference (\(r = -0.057\)).

\begin{table}[h!]
\centering
\begin{tabular}{lr}
\toprule
$$ & $target\_num$ \\
\midrule
$BFI2S_Open-Mindedness$  & $ 0.23$ \\
$BFI2XS_Open-Mindedness$  & $ 0.21$ \\
$BFI2S_Aesthetic Sensitivity$  & $ 0.18$ \\
$BFI2S_Intellectual Curiosity$  & $ 0.18$ \\
$BFI2S_Creative Imagination$  & $ 0.16$ \\
$BFI2S_Energy Level$  & $ 0.16$ \\
$MAIA_Not-Distracting$  & $ 0.14$ \\
$BFI2S_Agreeableness$  & $ 0.13$ \\
$BFI2S_Compassion$  & $ 0.12$ \\
$BFI2S_Trust$  & $ 0.11$ \\
$BFI2XS_Agreeableness$  & $ 0.07$ \\
$BFI2S_Respectfulness$  & $ 0.07$ \\
$BFI2S_Extraversion$  & $ 0.06$ \\
$BFI2S_Assertiveness$  & $ 0.06$ \\
$MAIA_Not-Worrying$  & $ 0.04$ \\
$BSI_Hostility$  & $ 0.04$ \\
$BSI_Anxiety$  & $ 0.02$ \\
$BSI_Somatization$  & $ 0.02$ \\
$MAIA_Emotional Awareness$  & $ 0.02$ \\
$MAIA_Attention Regulation$  & $ 0.01$ \\
$BSI_PST$  & $ 0.00$ \\
$MAIA_Self-Regulation$ &  $-0.04$ \\
$BFI2S_Depression$ &  $-0.05$ \\
$MAIA_Body Listening$ &  $-0.05$ \\
$BFI2XS_Extraversion$ &  $-0.06$ \\
$SWLS$ &  $-0.06$ \\
$BFI2S_Responsibility$ &  $-0.08$ \\
$BFI2S_Sociability$ &  $-0.09$ \\
$BSI_PDSI_temp$ &  $-0.10$ \\
$BSI_GSI$ &  $-0.10$ \\
$BSI_Interpersonal Sensitivity$ &  $-0.10$ \\
$BSI_Paranoid Ideation$ &  $-0.11$ \\
$BFI2S_Productiveness$ &  $-0.11$ \\
$BFI2S_Emotional Volatility$ &  $-0.12$ \\
$BSI_Phobic Anxiety$ &  $-0.12$ \\
$BSI_PDSI$ &  $-0.12$ \\
$MAIA_Noticing$ &  $-0.13$ \\
$BSI_Obsession-Compulsion$ &  $-0.13$ \\
$BFI2XS_Conscientiousness$ &  $-0.13$ \\
$BFI2S_Organization$ &  $-0.13$ \\
$BFI2S_Negative Emotionality$ &  $-0.14$ \\
$BFI2S_Conscientiousness$ &  $-0.14$ \\
$BSI_Psychoticism$ &  $-0.14$ \\
$BSI_Depression$ &  $-0.15$ \\
$BFI2XS_Negative Emotionality$ &  $-0.17$ \\
$MAIA_Trust$ &  $-0.17$ \\
$BFI2S_Anxiety$ &  $-0.20$ \\
\bottomrule
\end{tabular}

\caption{Pearson correlation (r) between aggregated one-time onboarding survey scores and the thermal comfort target.}
\label{table:correlation_pearson_onboarding_target}
\end{table}

\FloatBarrier
\subsection{Influence of metric selection}
\label{sec:metrics_results}

\begin{figure*}
  \centering
\begin{subfigure}{15cm}
    \includegraphics[width=0.9\linewidth]{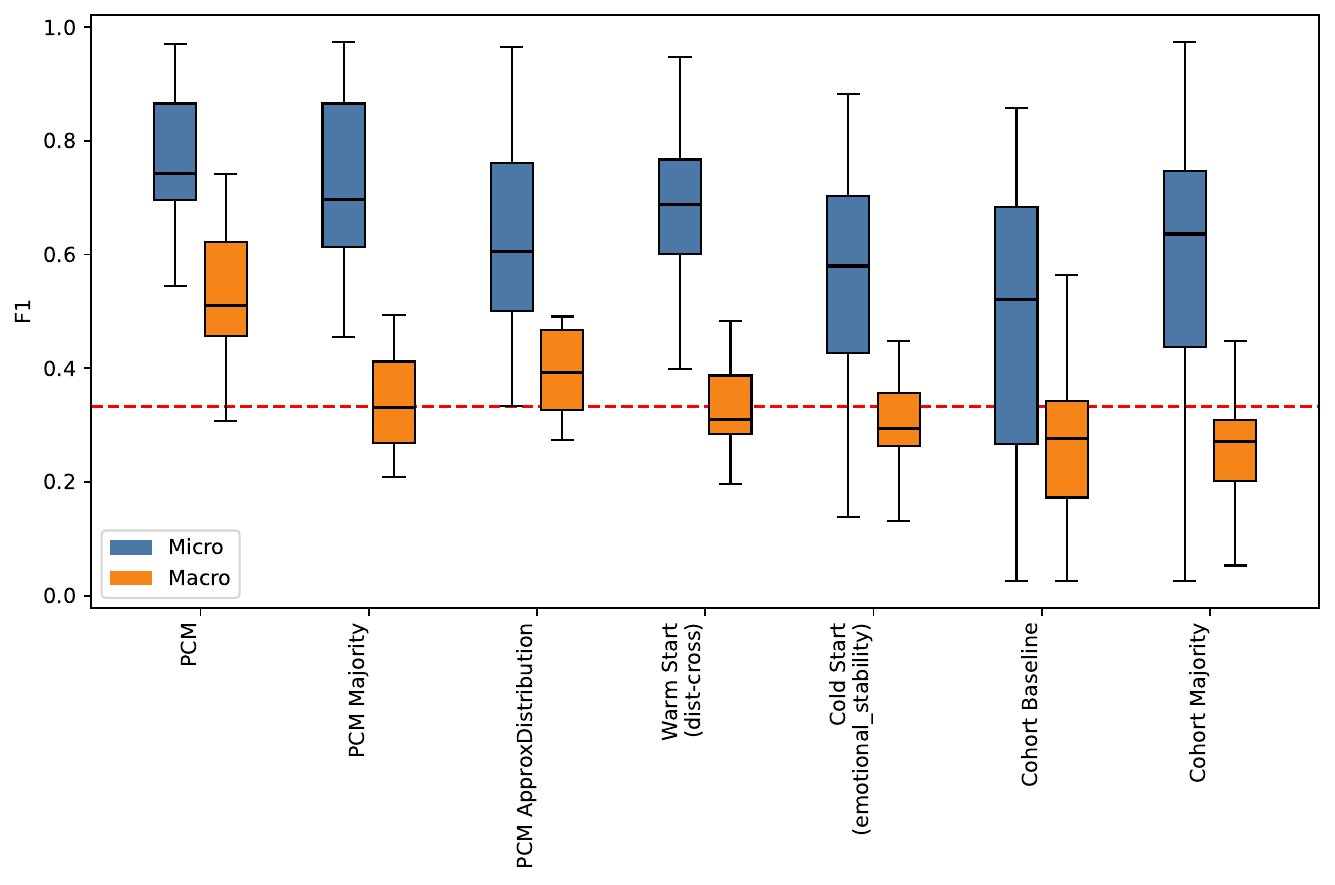}
 \caption{DORN} 
    \label{fig:dorn_micro_vs_macro_gap}
\end{subfigure}

\begin{subfigure}{15cm}
    \includegraphics[width=0.9\linewidth]{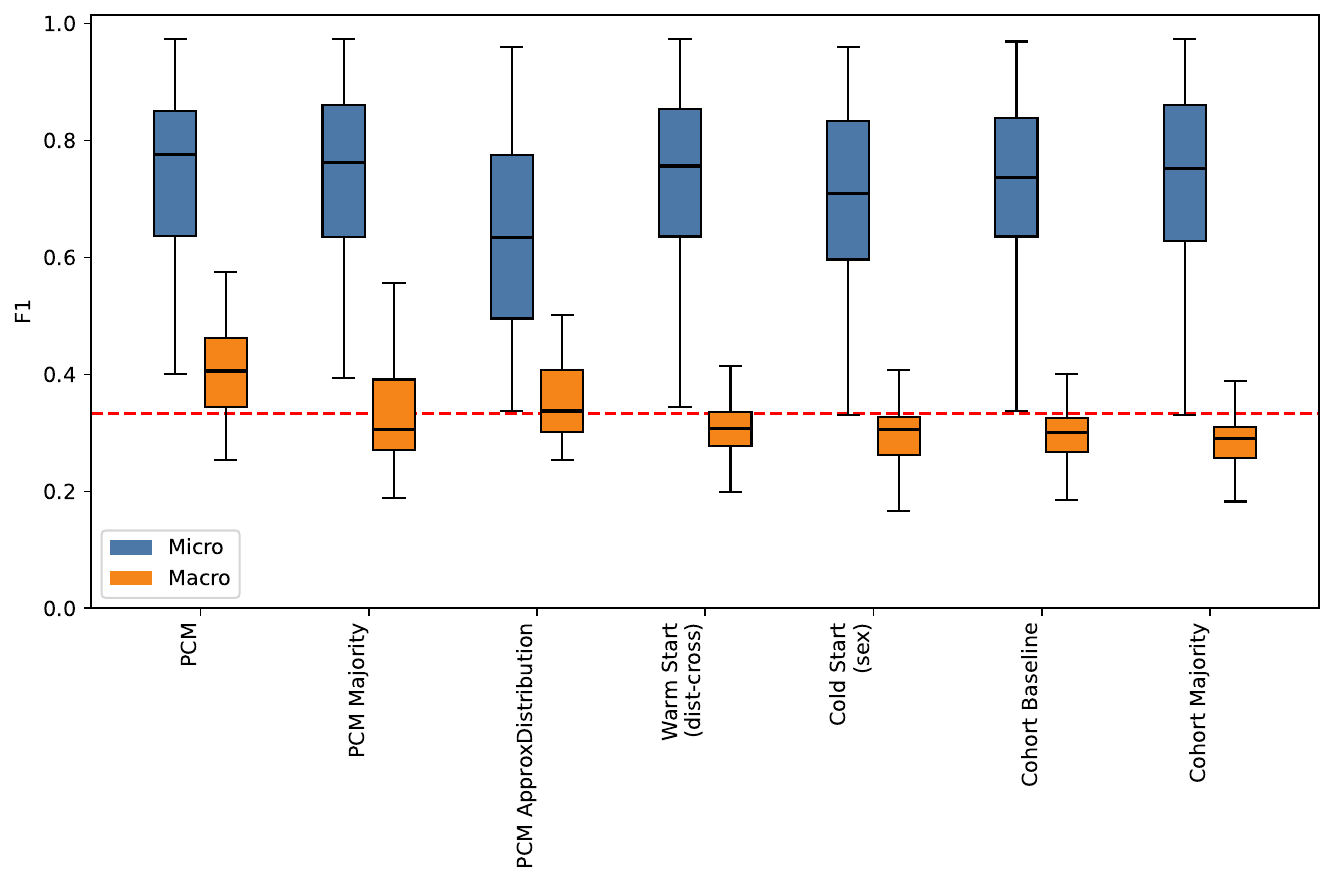}
 \caption{PersonAI}
    \label{fig:personai_micro_vs_macro_gap}
\end{subfigure}
\caption{Per-user distribution of F1-micro (blue) and F1-macro (orange) scores for the PCM, PCM Majority Baseline, PCM ApproxDist Baseline, best warm start cohort model, best cold start cohort model, cohort baseline, and the cohort majority baseline. The dashed red line signifies the Random 3-class F1 score.} 
\label{fig:micro_vs_macro}
\end{figure*}

Figures~\ref{fig:dorn_micro_vs_macro_gap} and~\ref{fig:personai_micro_vs_macro_gap} show the distributions of per-user F1-micro and F1-macro scores for the DORN and PersonAI datasets, respectively. 
The boxplots compare the PCM, the PCM Majority baseline, the ApproxDistribution baseline, the best-performing warm-start cohort model, the best-performing cold-start cohort model, cohort baseline (general-purpose model), and the cohort Majority baseline.
The PCM is a Random Forest classifier, the warm start cohort model uses cluster predictions from the combined response-distribution and cross-model performance approach (\texttt{Cross\_Dist\_survey}).
 
For the DORN dataset, the median F1-micro scores ranked as follows: PCM ($\approx 0.74$) $>$ PCM Majority ($\approx 0.70$) $>$ Warm Start ($\approx 0.69$) $>$ Cohort Majority ($\approx 0.64$)  $>$ ApproxDistribution ($\approx 0.61$) $>$ Cold Start ($\approx 0.58$) $>$ Cohort Baseline ($\approx 0.52$).
The median F1-macro scores ranked as follows: PCM ($\approx 0.51$ )$>$ ApproxDistribution ($\approx 0.39$) $>$  Majority ($\approx 0.33$)  $>$  Warm Start ($\approx 0.31$) $\approx$ Cold Start ($\approx 0.29$) $>$ Cohort Baseline ($\approx 0.28$) $>$ Cohort Majority ($\approx 0.27$).
The cohort-based models, median scores ranging from approximately 0.29 (cold) to 0.31 (warm) and the cohort baseline with 0.28, stay below the F1 macro score that is achieved by a uniformly random predictor on uniformly random data (0.33).

A similar pattern was observed for the PersonAI dataset (Figure~\ref{fig:personai_micro_vs_macro_gap}). 
The median F1-micro scores were closely clustered between approximately $0.71$ and $0.78$ across all models, including the Cohort and PCM Majority baselines. 
Using F1-macro, the Warm Start and Cold Start models perform worse than the Cohort basline (approximately $0.30$--$0.31$ vs.\ $0.32$), all of them staying below the F1 macro score and staying below the F1 score of a uniformly random predictor on uniformly random data (0.33), while the PCM achieved the highest median score (approximately $0.41$).

All methods, including the stoachastic or rule based baseline, achieve substantially higher F1-micro than F1-macro scores.

\FloatBarrier
\subsection{Classification performance}
\label{sec:classification_performance}
\begin{figure}[htbp]
    \centering
    \includegraphics[width=0.7\textwidth]{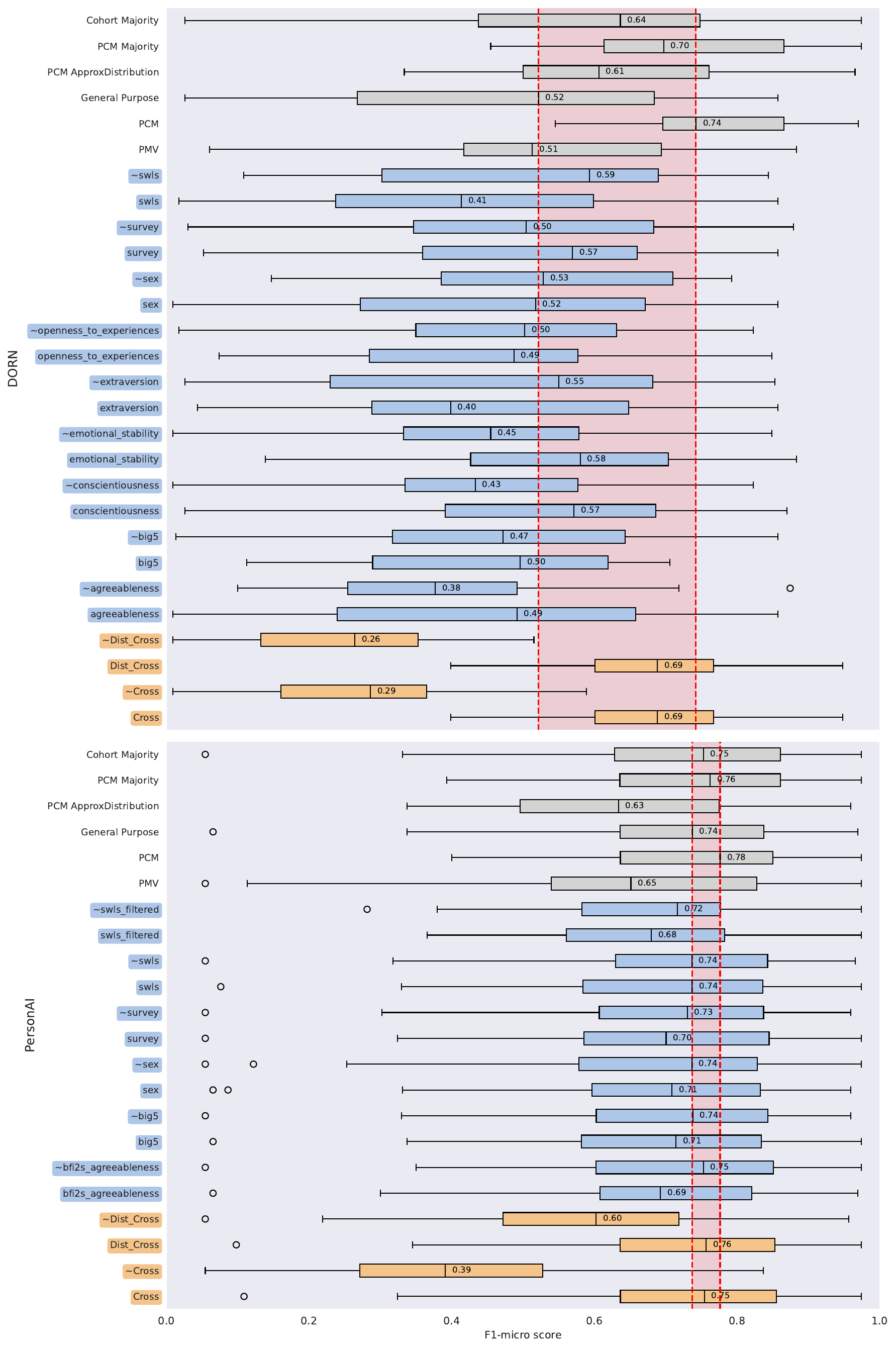}
    \caption{F1-micro performance comparison across all cohort approaches for the DORN (top) and PersonAI (bottom) datasets. Tilde-prefixed entries denote the ablation condition in which test users are purposely assigned to an incorrect cluster. The red band marks the range between the General Purpose and PCM medians. Cold and warm experiments are shown in blue and orange, respectively.
}    \label{fig:performance_comparison_fig4}
\end{figure}

\begin{figure}[htbp]
    \centering
    \includegraphics[width=0.7\textwidth]{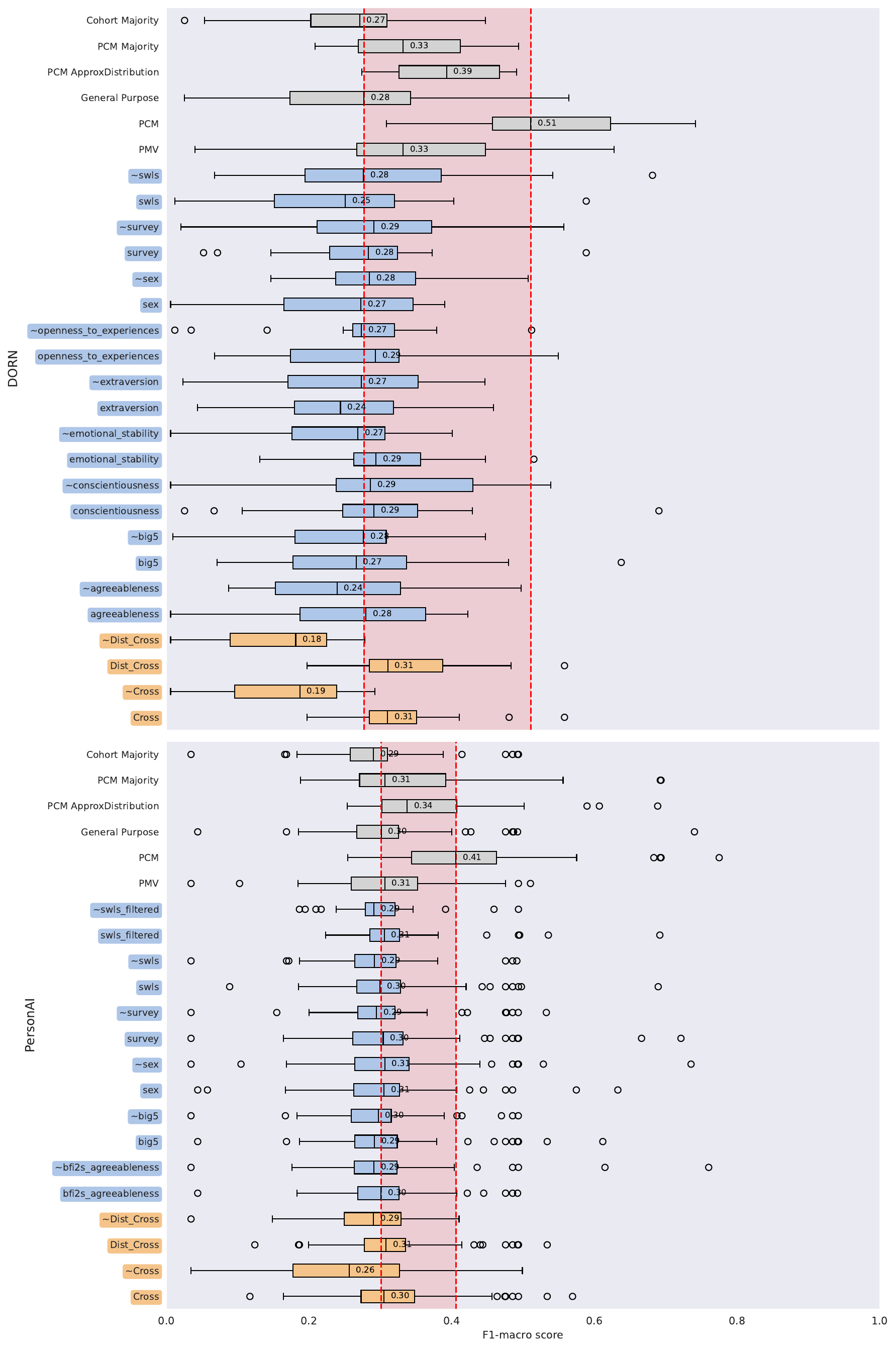}
    \caption{F1-macro performance comparison across all cohort approaches for the DORN (top) and PersonAI (bottom) datasets. Tilde-prefixed entries denote the ablation condition in which test users are purposely assigned to an incorrect cluster. The red band marks the range between the General Purpose and PCM medians. }
    \label{fig:performance_comparison_fig4_macro}
\end{figure}

The following methods were evaluated: the Personal Comfort Model (PCM), the General Purpose model, the Predicted Mean Vote (PMV) model, the Majority Class baselines, the per person random from approximate distribution model (PCM ApproxDistribution), the Cold Start cohort models, the Warm Start cohort models, and the Wrong Cluster controls.
Wrong Cluster controls assign participants to incorrect cohorts while keeping the underlying cohort models unchanged. Warm Start models use the ground-truth warm-start cohort assignments (PersonAI\_warm and DORN\_warm).
Figures \ref{fig:performance_comparison_fig4} and \ref{fig:performance_comparison_fig4_macro} show the F1-micro scores, computed per fold across all experiments. 
Each boxplot shows the score distribution across folds.

For the DORN dataset (Figure~\ref{fig:performance_comparison_fig4}, top), the median F1-micro scores were $0.52$ for the General Purpose model, $0.74$ for the PCM, $0.51$ for PMV, and $0.64$ for the Majority Class baseline. 
Among the cold start approaches the following cohorting methods contradict the expected direction of the ablation: \texttt{swls} ($0.41$ correct assigned vs.\ $0.59$ incorrect assigned), \texttt{extraversion} ($0.40$ correct vs. $0.55$ incorrect). 
Conversely, \texttt{survey} ($0.57$ vs.\ $0.50$), \texttt{emotional\_stability} ($0.58$ vs.\ $0.45$), \texttt{conscientiousness} ($0.57$ vs.\ $0.43$), and \texttt{agreeableness} ($0.49$ vs.\ $0.38$) all show the correct assignment outperforming incorrect assignment, while \texttt{sex} ($0.52$ vs.\ $0.53$) and \texttt{openness\_to\_experiences} ($0.49$ vs.\ $0.50$) show essentially no difference between correct and incorrect cluster assignment.
The warm start approaches show the largest correct-vs-incorrect gaps of any approach: \texttt{Cross} reaches $0.69$ (correct) vs.\ $0.26$ ($\sim$\texttt{Cross}), and \texttt{Dist\_Cross} reaches $0.69$ vs.\ $0.26$.
 
For the PersonAI dataset (Figure~\ref{fig:performance_comparison_fig4}, bottom), the median F1-micro scores were $0.74$ for the General Purpose model, $0.78$ for the PCM, and $0.75$ for the Majority Class baseline. 
The Majority Class baseline therefore performed similarly to both the General Purpose model and the PCM.
The Cold Start models showed little difference between correct and incorrect cluster assignment. 
For example, \texttt{sex} achieved median F1-micro scores of $0.71$ and $0.74$, while \texttt{agreeableness} showed the opposite pattern, with incorrect assignment outperforming correct assignment ($0.75$ vs.\ $0.69$). 
In contrast, the Warm Start models showed much larger differences: \texttt{Cross} achieved $0.75$ versus $0.39$, and \texttt{Dist\_Cross} achieved $0.76$ versus $0.60$.

When switching from F1-micro to F1-macro, many of the differences observed under F1-micro become much smaller.
For DORN (Figure~\ref{fig:performance_comparison_fig4_macro}, top), General Purpose drops to $0.28$, PCM to $0.51$, PMV to $0.33$, and Majority Class to $0.27$. 
All Cold Start models, regardless of whether cluster assignment was correct or incorrect, achieved median F1-macro scores between $0.24$ and $0.29$.
These values were similar to those of both the General Purpose model ($0.28$) and the Majority Class baseline ($0.27$).
The Warm Start models also outperformed the incorrectly assigned models under F1-macro, although the differences were smaller than under F1-micro. 
\texttt{Dist\_survey} achieved median F1-macro scores of $0.32$ and $0.16$ for the correct and incorrect assignments, respectively. The corresponding values for \texttt{Cross\_Dist\_survey} were $0.31$ and $0.18$.
All of the cohort models are performing below the F1 score produced by a uniformly random classifier on uniformly random data for a 3-class classification problem.
The PCM remained the best-performing model under both metrics.
Under F1-micro, the median scores were $0.74$ for PCM and $0.52$ for the General Purpose model; under F1-macro, the corresponding scores were $0.51$ and $0.28$.

For the PersonAI dataset (Figure~\ref{fig:performance_comparison_fig4_macro}, bottom), the General Purpose ($0.30$), PMV ($0.31$), and Majority Class ($0.29$) models achieved similar median F1-macro scores. 
All warm- and cold-start cohort models, with both correct and incorrect cluster assignments, also fell within a narrow range of $0.29$ to $0.31$. 
None exceeded the General Purpose model by more than 0.01 F1. 
All of the cohort models are performing below the F1 score produced by a uniformly random classifier on uniformly random data for a 3-class classification problem.
The PCM was the only model that clearly outperformed the others, with a median F1-macro score of $0.41$.
PCM models outperformed all other models in terms of F1 macro, followed by the PCM ApproxDistribution model on both datasets.

\FloatBarrier
\subsection{Class distribution}
\label{sec:class_distribution}

\begin{figure}[htbp]
    \centering
    \includegraphics[width=0.5\textwidth]{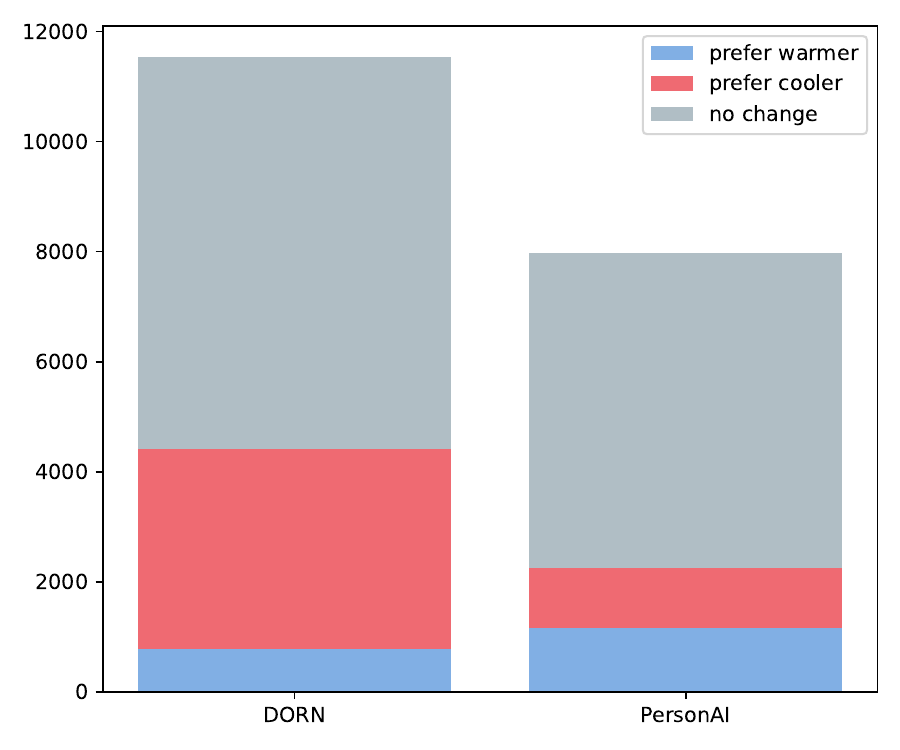}
   \caption{Class distribution of thermal preference labels for the DORN and PersonAI datasets. Both datasets are dominated by the ``no change'' class.}    \label{fig:class_distribution_per_dataset}
\end{figure}

\begin{figure}
	\centering
	\begin{subfigure}{0.45\textwidth}
    \centering
    \includegraphics[width=\textwidth]{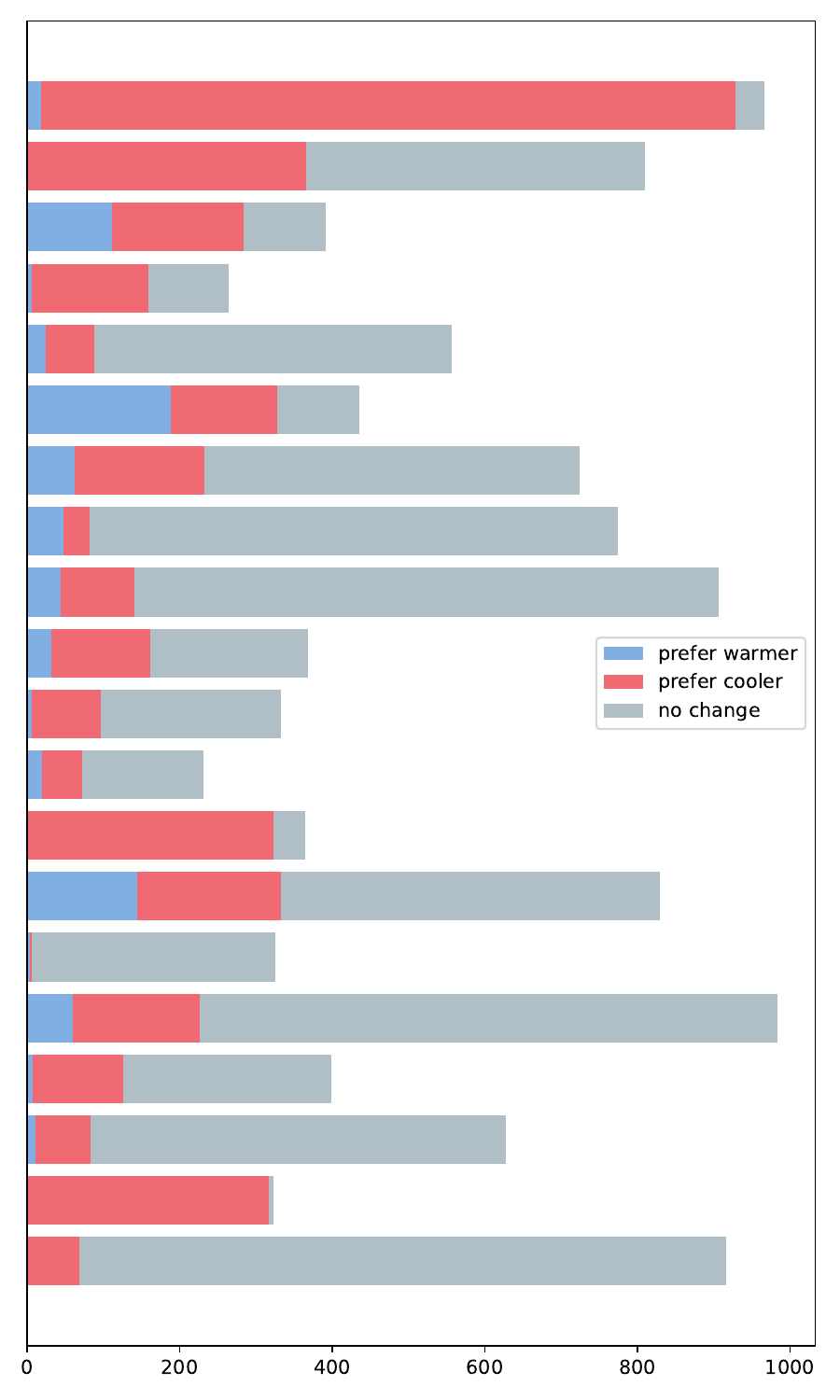}
   \caption{DORN}    
   \label{fig:class_distribution_per_person_dorn}
\end{subfigure}
    \hfill
	\begin{subfigure}{0.45\textwidth}
    \centering
    \includegraphics[width=\textwidth]{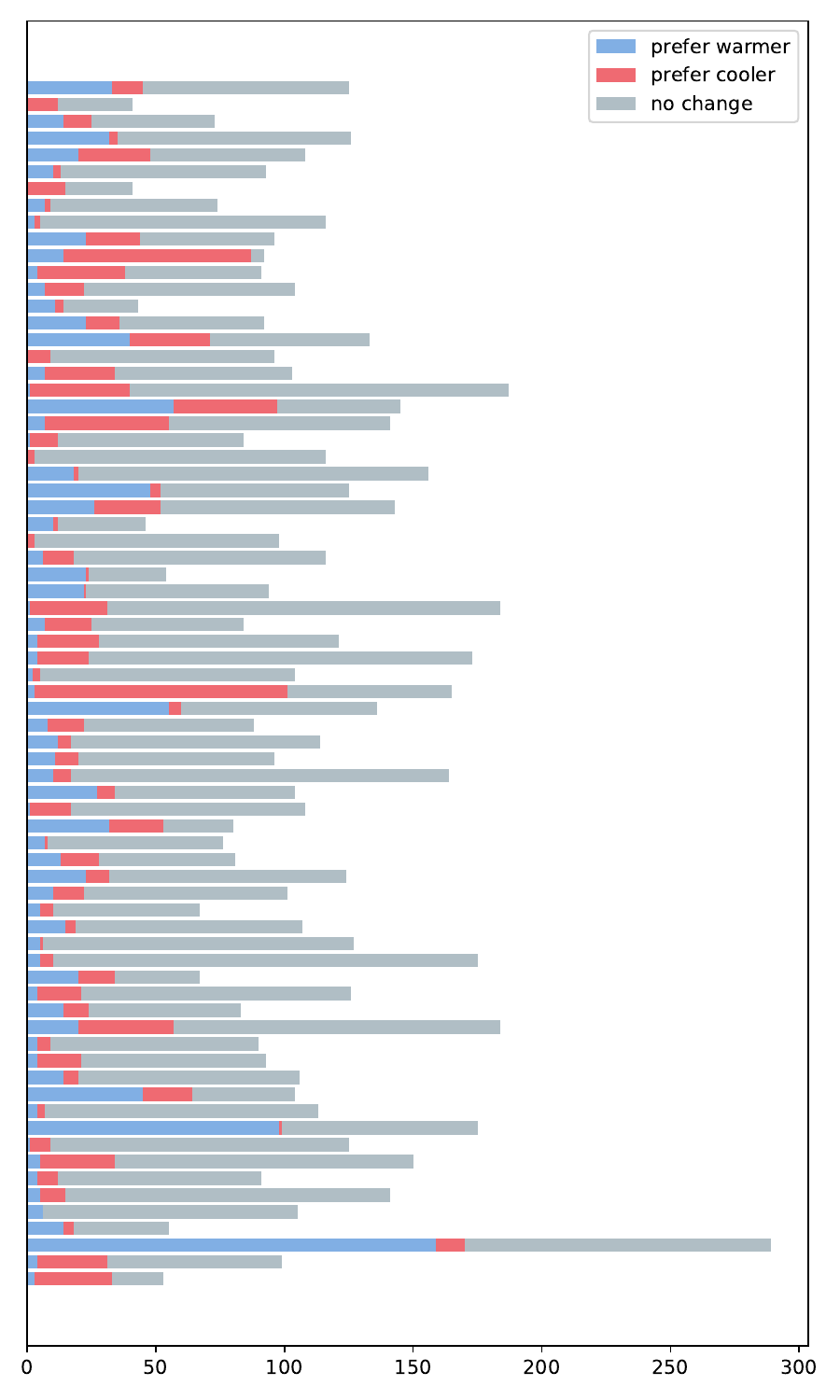}
   \caption{PersonAI}    
   \label{fig:class_distribution_per_person_personai}
\end{subfigure}
	\label{fig:class_distribution}
     \caption{Class distribution of thermal preference labels per participant.}    
\end{figure}

The differences between F1-micro and F1-macro reflect the class imbalance in the two datasets. 
As shown in Figure~\ref{fig:class_distribution_per_dataset}, the ``no change'' class dominates both DORN and PersonAI, whereas the ``prefer cooler'' and ``prefer warmer'' classes are much less frequent. 
The class distribution on a participant basis is shown in Figure~\ref{fig:class_distribution_per_person_dorn} for the DORN dataset and in Figure~\ref{fig:class_distribution_per_person_personai} for the PersonAI dataset.
This asymmetry is precisely the condition under which F1-micro, being equivalent to overall accuracy in a single-label multi-class setting, rewards models for defaulting to the majority class and penalizes minority-class errors only marginally.

\begin{figure}
	\centering
	\begin{subfigure}{0.45\textwidth}
    	\centering
		\includegraphics[width=\textwidth]{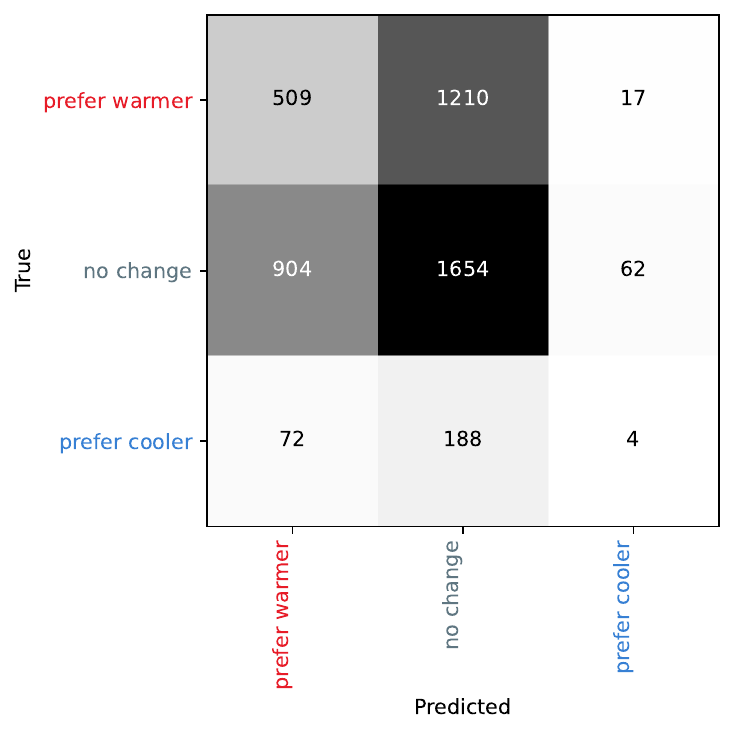}
   \caption{Confusion matrix of the general-purpose (cohort baseline) model on the full DORN dataset. } 
\label{fig:dorn_confusion_matrix_baseline_cohort_full}
\end{subfigure}
    \hfill
	\begin{subfigure}{0.45\textwidth}
         \centering
		\includegraphics[width=\textwidth]{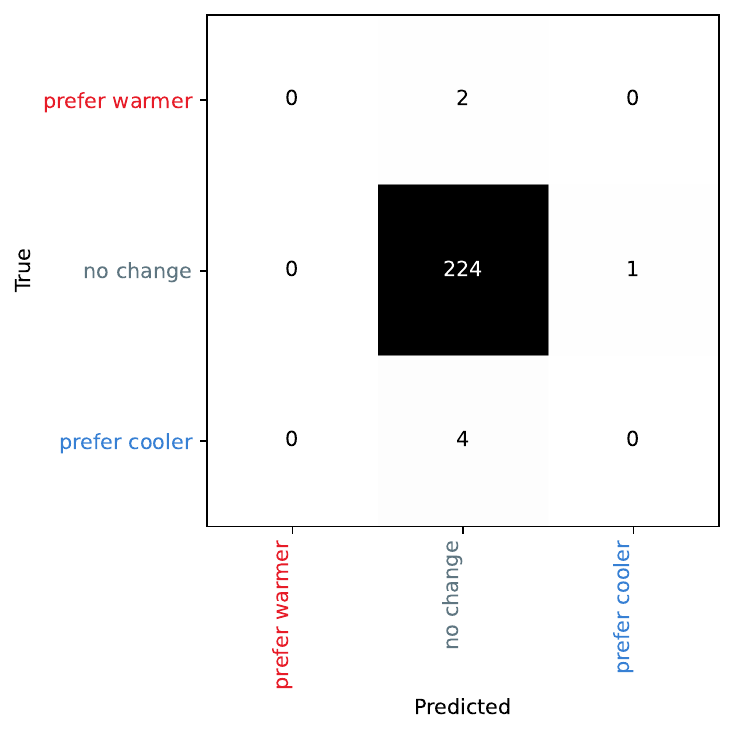}
    \caption{Confusion matrix of the PCM for user \texttt{dorn14} in the DORN dataset, an example of an extremely imbalanced individual user.}  
\label{fig:dorn_confusion_matrix_pcm_imbalanced_user_dorn14}	\end{subfigure}
	\label{fig:confusion_matrices}
\end{figure}

These imbalances lead to a classification problem that is highly skewed, as shown in the confusion matrix of the general-purpose / baseline model on the DORN dataset (Figure~\ref{fig:dorn_confusion_matrix_baseline_cohort_full}). 
Aggregated across all users, the baseline model achieves an overall accuracy (equivalent to F1-micro) of $2167/4620 \approx 46.9\%$, but this figure masks a strongly class-dependent behavior: recall for ``no change'' is $1654/2620 \approx 63.1\%$, recall for ``prefer warmer'' is $509/1736 \approx 29.3\%$, and recall for ``prefer cooler'' collapses to just $4/264 \approx 1.5\%$. 
The model is heavily biased toward predicting ``no change'' and ``prefer warmer'' and almost never predicts ``prefer cooler'' correctly, even though that class accounts for a non-trivial share of the ground truth, illustrating the class-dependent failure mode hidden by an aggregate F1-micro score. 
This class-wise disparity is not well captured by F1-micro, but is reflected by F1-macro, which averages performance equally across classes.

However, this problem is even more pronounced on the PCM level.
Many users have highly imbalanced thermal preference distributions, while making their responses easy to predict and inflating the overall performance scores.
Figure~\ref{fig:dorn_confusion_matrix_pcm_imbalanced_user_dorn14} shows an example PCM for such a user in the DORN dataset (user \texttt{dorn14}): of the 231 labeled instances for this user, 225 are ``no change,'' with only 2 ``prefer warmer'' and 4 ``prefer cooler'' instances. 
The PCM predicts almost exclusively ``no change,'' achieving an accuracy of $224/231 \approx 97.0\%$ (F1-micro) while failing to correctly predict any instance of either minority class (0\% recall for both ``prefer warmer'' and ``prefer cooler''). This user alone would contribute a near-perfect F1-micro score, despite the model failing completely on the minority classes.

\FloatBarrier
\subsection{Participant-level performance changes}
\label{sec:outliers}

\begin{figure}
	\centering
	\begin{subfigure}{\textwidth}
    \includegraphics[width=\textwidth]{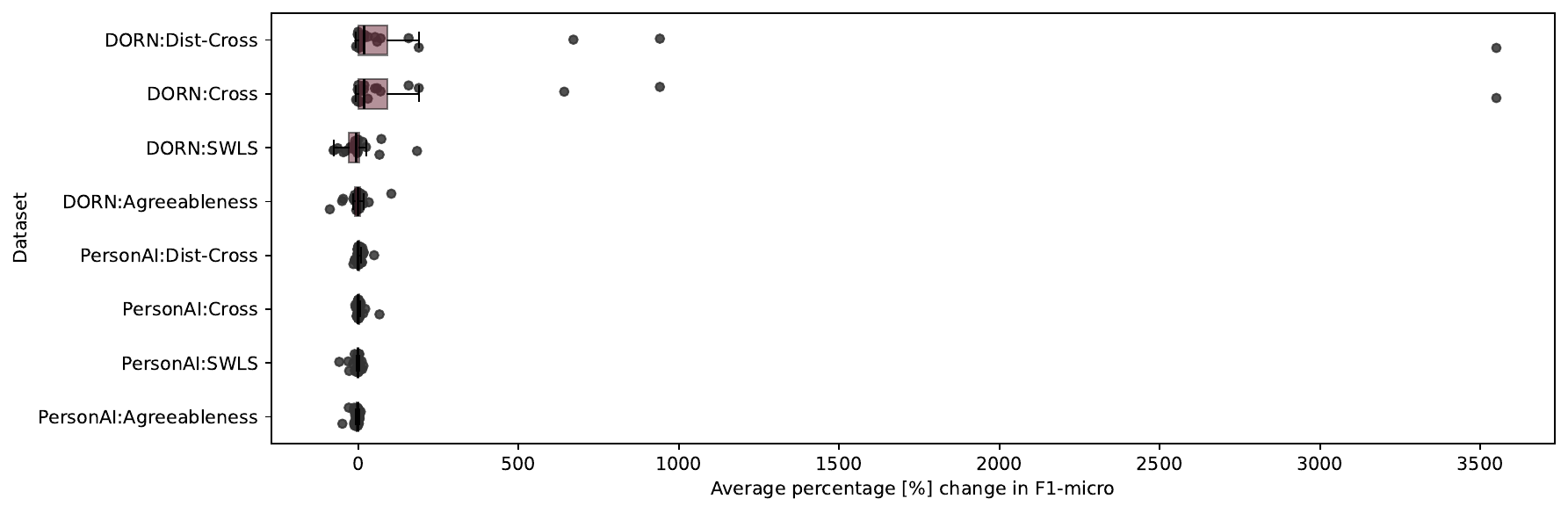}
   \caption{Full range.}    \label{fig:point_clouds_personai_dorn}
\end{subfigure}
    \hfill
	\begin{subfigure}{\textwidth}
    \centering
    \includegraphics[width=\textwidth]{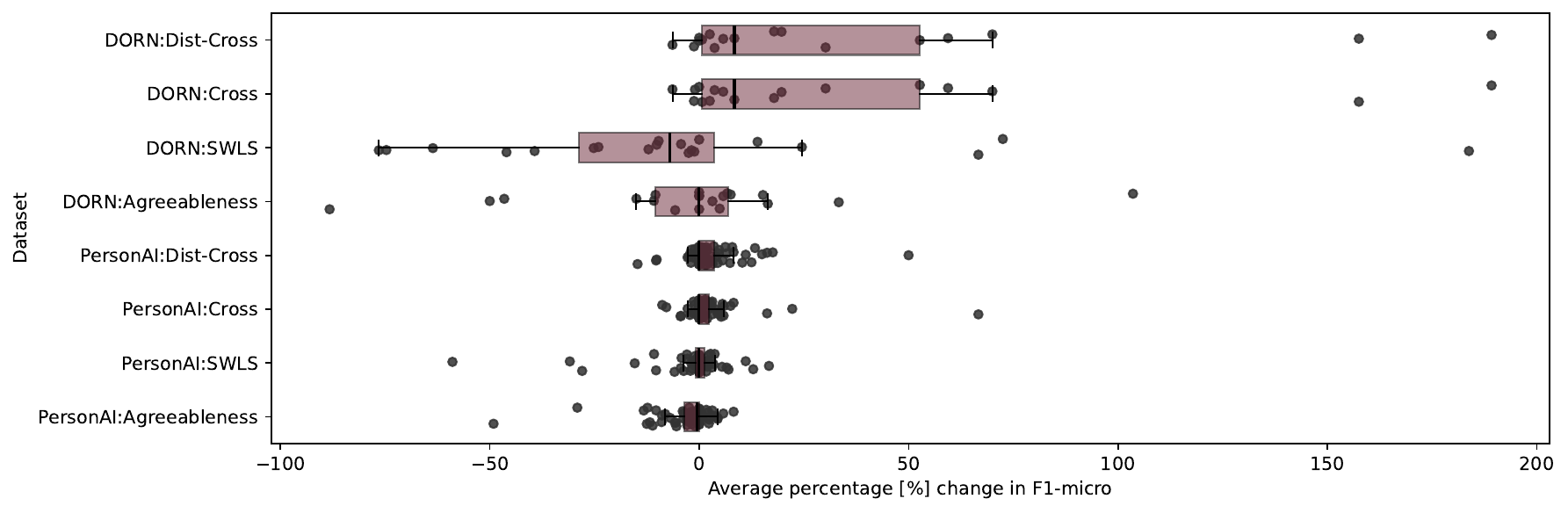}
    \caption{Restricted to gaps under $200\%$.}    \label{fig:point_clouds_personai_dorn_limited200}
\end{subfigure}
   \caption{Per-user percentage change in F1-micro from the general-purpose model to each cohort/PCM approach, for DORN and PersonAI.}  
	\label{fig:point_clouds}
\end{figure}

To examine the drivers behind the observed performance increases of specific cohort modes, we replicated the per-user percentage change in performance from the General Purpose model to the cohort models using the methodology of the original study and applying F1 micro as metric. Figure~\ref{fig:point_clouds_personai_dorn} shows the per-user percentage change in F1-micro,$\Delta = 100 \times (\mathrm{coh} - \mathrm{base})/\mathrm{base}$ for the \texttt{Agreeableness}, \texttt{SWLS}, \texttt{Dist}, and \texttt{Dist-Cross} approaches on both datasets.
On the DORN dataset, a small number of users showed percentage improvements exceeding $500\%$, with the largest values reaching approximately $3500$--$3600\%$ for the \texttt{Dist} and \texttt{Dist-Cross} approaches. 
The PersonAI dataset showed the same pattern, although the largest percentage improvements were substantially smaller (up to approximately $190\%$).
These extreme percentage improvements arose because some users had very low baseline F1-micro scores. 
As illustrated in Figure~\ref{fig:dorn_confusion_matrix_pcm_imbalanced_user_dorn14}, these users typically had highly imbalanced thermal preference distributions. Consequently, even small absolute improvements produced very large percentage changes. 
Overall, the reported improvement in cohort model performance was driven by a small number of users with very large relative gains.

Figure~\ref{fig:point_clouds_personai_dorn_limited200} shows the same comparison after restricting the percentage change to values below $200\%$, making the distribution for most users easier to inspect. 
For the DORN dataset, the median improvement over the General Purpose model was approximately $8\%$ for both the \texttt{Dist-Cross} and \texttt{Dist} approaches. 
The corresponding interquartile ranges were approximately $0$--$50\%$ and $0$--$30\%$, respectively.
The \texttt{SWLS} and \texttt{Agreeableness} approaches were centred close to $0\%$, and several users experienced decreases of $75$--$90\%$ relative to the General Purpose model. 
For the PersonAI dataset, all four approaches were centred close to $0\%$, with most users showing changes within approximately $\pm20\%$ and only a few reaching $150$--$190\%$.
Overall, these results indicate that the reported micro averaged F1 improvement of the cohort framework is driven by a small number of users with very large relative gains, whereas the typical user experiences little or no improvement over the General Purpose model.

\FloatBarrier
\subsection{Dataset Size}
\label{sec:dataset_size}

\begin{figure}
	\centering
	\begin{subfigure}{0.45\textwidth}
    	\centering
		\includegraphics[width=\textwidth]{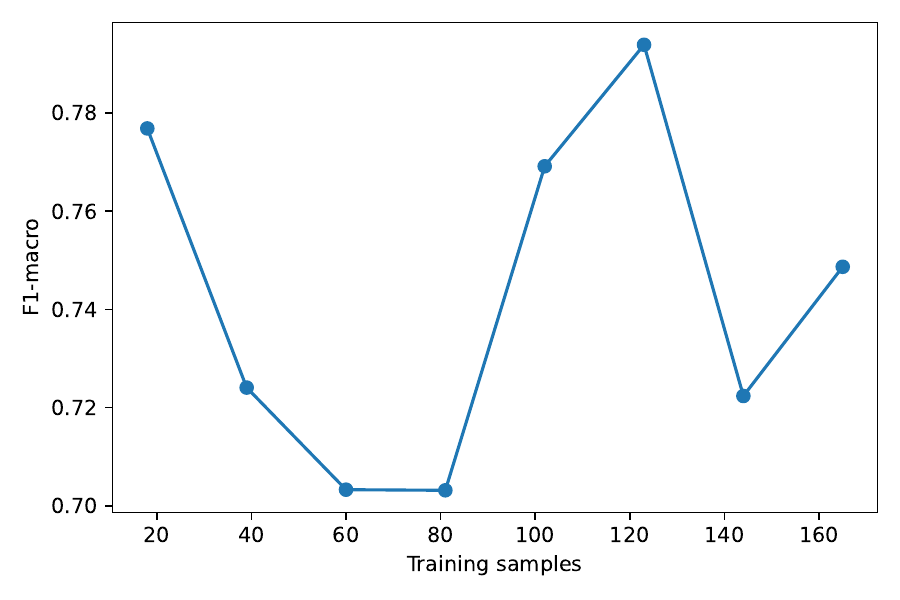}
    \caption{User 001 from DORN Dataset}
\label{fig:learning_curve_dorn1}
\end{subfigure}
    \hfill
	\begin{subfigure}{0.45\textwidth}
         \centering
		\includegraphics[width=\textwidth]{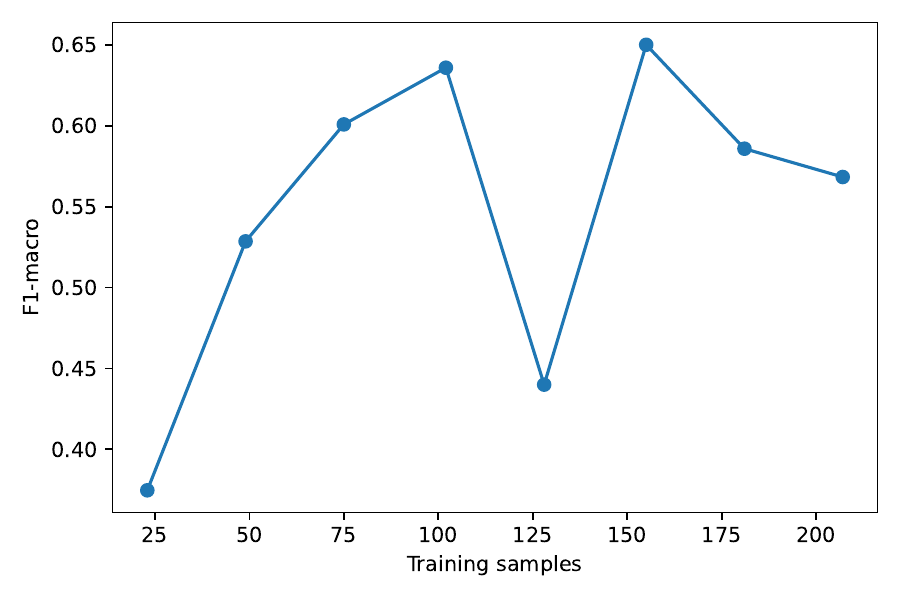}
    \caption{User 003 from PersonAI dataset} 
\label{fig:learning_curve_personai3}
\end{subfigure}
	\label{fig:learning_curves}
        \caption{Learning curves for the participants with the most available data for PCM models in each dataset} 
\end{figure}

The learning curves of the PCM models for both the DORN and PersonAI datasets show no evidence of convergence or a consistent performance trend. 
Figures~\ref{fig:learning_curve_dorn1} and~\ref{fig:learning_curve_personai3} show the learning curves for the participants with the largest number of observations in each dataset.

The average number of available datapoints per person in the DORN dataset is 567, for the PersonAI dataset this number is lower at 110. The average size of the minority class per person in DORN and PersonAI dataset is 34.3 and respectively 6.8.
There are 3 participants in the DORN dataset that have no datapoints in at least one class, the same is true for 6 participants in the PersonAI dataset.
There are 5 participants in DORN dataset and 38 participants in PersonAI dataset for which there are less than 5 datapoints available for the minority class.

\section{Discussion}
The authors acknowledge that many of the methodological issues identified in this study could, in principle, have been detected through a careful methodological review of the original work. 
However, these issues only became apparent during the process of reproducing the study and analysing the PersonAI own dataset. 
This highlights one of the strengths of independent replication studies bringing a broader range of expertise to the subject. Beyond assessing reproducibility, they can reveal methodological limitations and practices that may not be apparent from the published work alone, while also providing insight into the research context in which the original study was conducted.

\subsection{Too little data}
\label{sec:too_little_data}
Quintana et al.\ \cite{quintanaCohortComfortModels2023} justify the amount of training data required for personal thermal comfort modelling by referring to several prior studies \cite{kimPersonalComfortModels2018, liPersonalized2017, liuPersonalThermalComfort2019}. 
Our own results (Section~\ref{sec:dataset_size}, Figs.~\ref{fig:learning_curve_dorn1} and \ref{fig:learning_curve_personai3}) show no clear convergence even for the participants with the most available data, which calls into question the widely repeated assumption that around 60 data points per person is sufficient for stable thermal preference prediction. 
The other two studies commonly cited alongside it also rely on small samples.
Li et al.~\cite{liPersonalized2017} base their conclusions on data from only three participants (271 observations in total), yet conclude that ``with a relatively small dataset (approximately 50 samples), the comfort model can achieve an acceptable classification accuracy.''
Liu et al. \cite{liuPersonalThermalComfort2019} study 14 participants with an average of 275 data points each, follow a very similar experimental and evaluation style to Kim et al., and attribute some of their own models' low predictive power to participants not "answering carefully" or being inherently less predictable.
Notably, using standard interpretation guidelines for Cohen's kappa (0.01--0.20 as none to slight, 0.21--0.40 as fair, 0.41--0.60 as moderate, 0.61--0.80 as substantial, and 0.81--1.00 as almost perfect agreement), their reported best-case kappa of 0.25 would itself be classified as only "fair" agreement, which is a further indication that the underlying models are not capturing the target well even in the most favorable case reported.

Since Kim et al.~\cite{kimPersonalComfortModels2018} is the primary source of the widely cited estimate of 60 data points per person, we examine the basis for this estimate in more detail.
The paper introduces the concept of ``first convergence,'' obtained by training the same model on increasingly large training set sizes and producing a learning curve for each participant (their Fig.~A1). 
``First convergence'' is defined as the point at which the derivative of the learning curve plateaus (within $\pm 0.001$) for two successive runs. 
The notion of ``first convergence'' is therefore a heuristic criterion that is in no way related to the mathematical concept of ''convergence'', or the ML concept of ``model convergence``, and is unsuitable to serve as estimate of how much training data is required.
The ``60 data points'' figure is obtained by averaging the training set size at which this first plateau occurs across all participants.
The following statement from Kim et al.~\cite{kimPersonalComfortModels2018} is particularly noteworthy: ``Note that some models go through more than one convergence. 
This is because the model has to relearn new patterns in the data with the addition of new data points.''
Given the pronounced class imbalance in the dataset, the observed behaviour is also consistent with a model learning the marginal class distribution rather than the relationship between the predictors and an individual's thermal preference. Small-sample regimes are known to carry a substantial risk of overfitting and optimistic performance estimates \cite{vabalasMachineLearningAlgorithm2019}.
The sample size required to train and evaluate a classifier depends on many factors, including the strength of the predictive signal, class balance, model complexity, feature dimensionality, and the dependence between observations.
To provide an estimate of dataset sizes required we consider:

\textbf{Class availability}
In order to apply subervised ML classification labeled training data has to be available. In order to train a classifier to distinguish 3 classes, all 3 classes must occur in the traing data. This is not the case for 15\% of the participants of the DORN dataset (3 out of 20), and  9\% of the participants in the PersonAI dataset (8 out of 66) (see Section\ref{sec:dataset_size}).
In order to measure and evaluate the performance of a ML based classification approach, the test set also must contain all classes.
Given that 5-fold-CV is used to train the PCM models, this gives us a theoretical limit of at least 5 datapoints per class to be available for each participant. This limit is not met for 25\% of participants in DORN dataset, and 52\% of participants in the PersonAI dataset.
Per class \emph{Recall}, and in consequence per class F1, may be undefined for test data that is missing this class. In practice, ML software libraries use a class F1 of 0 in this case.
In order to properly assess classification performance of a model, there must be sufficient test data available for every class to establish a desired resolution of performance metrics and expressiveness of the experiment.
For example, given 20 data points for a single class in a test set, allows the calculation of the classes \emph{Recall} in a resolution of 0.05 increments. For more than the half of DORN participants this resolution is not achievable.

\textbf{EPV}
Events per variable (EPV) originate from logistic regression and since has been used in a more general ML context for estimating a lower bounds for dataset sizes.
EPV is calculated by dividing the number of datapoints of the minority class by the number of predictors.
The corresponding rule of thumb is at least "10 events per variable"\cite{peduzziSimulationStudyNumber1996}.
The feature set used for training PCM on the DORN dataset consists of 5 features (Air temperature, Relative humidity, Near-body temperature, Heart rate, Clothing level).
Performing this calculation, shows us that at least 50 data points per class are required to achieve an $EPV<=10$, the mean per person minority class is 34.3 in the DORN dataset, and 6.8 in the PersonAI dataset.
Only 25\% of participants in the DORN dataset would exceed this limit.
PersonAI uses 6 features in the PCM experiments, providing a lower bound of 60 datapoints per class to satisfy the "10 events per variable rule". 
None of the participants in the PersonAI dataset satistfy this condition.
While some researchers critique the "10 events per variable rule" as an oversimplification that under certain circumstances lower EPV can be mitigated \cite{vansmedenNoRationale12016}, other reseachers have found that for approaches other than logistic regression EPV should be greater than 100 \cite{vanderploegModernModellingTechniques2014}, putting the number of datapoints per minority class around 500-600 for the PCM models.

\textbf{Related work on dataset sizes}
Previous research into data set size requirements for ML classification of medical and psycological data show, depending on the task at hand, a minimum of few hundreds of datapoints per class for a workable solution, and thousands of points per class to achieve good classification performance\cite{althnianImpactDatasetSize2021, rajputEvaluationDecidedSample2023, sordoSampleSizeClassification2005, EffectDatasetSize2021, beleitesSampleSizePlanning2013}.
To emphasize, the lowest starting point in any of the above cited analysis is a dataset size of 100 items.

Regarding the utilization of existing research datasets, Riley et al. \cite{rileyImportanceSampleSize2025} summarized:
"Many AI research studies rely on existing datasets that are readily available rather than recruiting new participants, as done in randomised trials or prospective cohort studies. Using an existing dataset provides a convenience sample, but researchers should not proceed before considering whether the sample size is adequate. When the existing dataset is considerably small, abstaining from using it, and instead seeking additional data is often a more appropriate course of action."

\textbf{Oversampling}
There are well researched methods for generating additional minority class datapoints, such as SMOTE \cite{chawlaSMOTESyntheticMinority2002} and ADASYN \cite{heADASYNAdaptiveSynthetic2008}
These methods however are data-driven methods that require a certain amount of data to start with, and must not be applied to the test data.
There is too little available data in both the PersonAI and DORN datasets to fulfil those criterea.
We did not correct this issue in our reproduction, and instead used the DORN dataset with the same per-person sample sizes (261) as was done by Quintana et al.\ \cite{quintanaCohortComfortModels2023}'s experiments.

\subsection{Distribution of predictions}
\label{sec:prediction_distribution}

While the class imbalance present in the original datasets is documented in both Quintana et al. \cite{quintanaCohortComfortModels2023} and Kim et al. \cite{kimPersonalComfortModels2018}, the distribution of predictions actually produced by their models is not. These predictions form the basis of every reported performance score, yet neither paper reports or visualizes them.
One established way of examining classifier behaviour beyond a single aggregate score is the confusion matrix. Producing confusion matrices would have revealed that the models are heavily biased toward the majority class, and that for many individuals the resulting PCMs effectively reduce to a two-class, or even one-class, classification problem rather than the three-class problem the framework sets out to solve. 
We addressed this gap by producing confusion matrices for both datasets and including them in this paper to make the models' actual predictive behaviour transparent (Figs.~\ref{fig:dorn_confusion_matrix_baseline_cohort_full} and \ref{fig:dorn_confusion_matrix_pcm_imbalanced_user_dorn14}).

\subsection{Use of improper metrics}
\label{sec:improper_metrics}
The use of micro-averaged F1 on imbalanced data suffers from the same well-documented issues as accuracy: in a single-label multi-class setting, micro-F1 is mathematically equivalent to overall accuracy, and both are known to be misleading measures of classifier performance under class imbalance \cite{grandiniMetricsMultiClassClassification2020,opitzCloserLookClassification2024, japkowiczEvaluatingLearningAlgorithms2011}. 
The resulting metric shows how well a model fits the label distribution of a test set, not how well it can discriminate between classes. 
Small improvements in micro-F1 may therefore reflect better majority-class prediction rather than any meaningful gain in detecting minority preferences, exactly the pattern we observe in Section~\ref{sec:classification_performance} and Section~\ref{sec:class_distribution}, where the trivial majority-class baseline is matches or outperforms most cohort approaches under F1-micro (Fig.~\ref{fig:performance_comparison_fig4}), and where per-class recall for the minority ``prefer cooler'' class collapses to under 2\% even for the general-purpose model (Fig.~\ref{fig:dorn_confusion_matrix_baseline_cohort_full}). 
Complementary, class-balanced metrics are therefore necessary to fully assess model performance.
Macro-averaged F1 addresses this to some degree \cite{grandiniMetricsMultiClassClassification2020,opitzCloserLookClassification2024, japkowiczEvaluatingLearningAlgorithms2011, heLearningImbalancedData2009} by weighting all classes equally regardless of their support, and is the metric under which the apparent gains of the cohort and warm-start approaches largely disappear in our reanalysis (Section~\ref{sec:classification_performance}, Fig.~\ref{fig:performance_comparison_fig4_macro}). 
However, for data as imbalanced as ours, where individual users may contribute few or even no instances of a given class within a fold (Section~\ref{sec:class_distribution}, Fig.~\ref{fig:dorn_confusion_matrix_pcm_imbalanced_user_dorn14}), the interpretation of macro-F1 also requires caution because precision and recall are undefined for absent classes.
This impacts the experiments and evaluation of the PCM models, as well as the evaluation of the cohorting models in a leave one out manner on user basis - generating per class F1 scores of 0 for the unavailable classes.
We added macro-averaged F1 scores in our reproduction study to provide a more complete picture of model performance alongside the originally reported micro-F1 scores. 
However, as discussed above, macro-F1 also has important limitations and should not be considered a standalone solution for evaluating classifiers on imbalanced data \cite{heLearningImbalancedData2009,japkowiczEvaluatingLearningAlgorithms2011}. 
Rather, it should be viewed as one of several complementary metrics for assessing per-user thermal preference models.
The macro ROC-AUC metric utilized by Kim et al. \cite{kimPersonalComfortModels2018}, being macro averaged, is another suitable metric for such imbalanced classification problems.

\subsection{Missing trivial baselines}
\label{sec:missing_baselines}
Baselines are an essential safeguard against illusory progress. 
A new method is only convincing if it clearly outperforms simple baselines, such as majority-class prediction and a random noise baseline for classification tasks as the one at hand.
Without these comparisons, apparent improvements may reflect properties of the data or the evaluation metric rather than a genuine improvement.
Given the task of multiclass classification, it is important to understand the used metrics, their limitations, and the performance envelope one aims to operate in.
Important considerations that form this performance envelope in our experiment are the F1 score of a uniformly random predictor applied to uniformly random data.
While such a random baseline is provided in the work of Kim et al. \cite{kimPersonalComfortModels2018}, it is missing in Quintana et al. \cite{quintanaCohortComfortModels2023}.
Further, given such imbalanced data, a majority class predictor and a random predictor adhereing to the target distribution may be considered.
Without such baselines, it is difficult to determine whether the reported improvements reflect genuine personalization or simply the strong class imbalance present in the data.
Further, such baselines can help finding errors and problems in the experiments in an early stage.
Our reproduction addressed this by including a majority-class baseline, a uniformly random predictor on uniformly random target, and a random predictor adhering to the target distribution, to provide context for the reported performance gains.
In our results we show that the F1 micro performance of the majority-class baseline outperforms all cohorting model approaches on the DORN dataset and performs about equally well to them in the PersonAI dataset (see Section \ref{sec:classification_performance}, Figure \ref{fig:performance_comparison_fig4}).
Which is a strong indicator that the utilized metric is not a good indicator for measuring the capabilities of a classifier to distinguish the three classes.
We show that the F1 macro performance of all cohorting models in both datasets is below the performance of the uniformly random predictor with 0.33 (see Section \ref{sec:classification_performance}, Figure \ref{fig:dorn_micro_vs_macro_gap} and Figure \ref{fig:personai_micro_vs_macro_gap}).
This is a result of the trained models overfitting to the majority class and either never predicting at least one class, or at least one class often being missing from the test set.
We show that the F1 macro performance of all cohorting models in both datasets is below the performance of the target distribution adhereing random classifier (see Section \ref{sec:classification_performance}, Figure \ref{fig:dorn_micro_vs_macro_gap} and Figure \ref{fig:personai_micro_vs_macro_gap}).
This demonstrates that the features used in both the models themselves, as well as the features used for cohort creation and model selection, do not infuse more information that the models can leverage to produce better predictions - since a completely feature oblivous random actor outperforms them.

\subsection{Methodological limitations of the experimental setup}
\label{sec:weak_experiment_setup}

Quintana et al. \cite{quintanaCohortComfortModels2023} perform 100 randomized repetitions of 80/20 train-test splits, while perform 150 randomized repetitions of 2-fold cross-validation. 
This so-called ``repeated cross-validation'' or ``Monte Carlo cross-validation'' has been shown to be statistically weaker than regular $k$-fold CV while being very computationally expensive \cite{yatesCrossValidationModel2023, cawleyOverfittingModelSelection2010, Vanwinckelen2012, michelucciFundamentalMathematicalConcepts2024, nakatsuValidationMachineLearning2023, burmanComparativeStudyOrdinary1989}, with some authors calling it ``a waste of resource'' \cite{Vanwinckelen2012}. 
There is a general consensus that leave-one-out cross-validation is to be preferred where computationally feasible \cite{burmanComparativeStudyOrdinary1989, yatesCrossValidationModel2023, cawleyOverfittingModelSelection2010}, and it has even been referred to as the ``gold standard'' \cite{yatesCrossValidationModel2023} among cross-validation methods. 
Such resampling methodologies (the various flavors of cross-validation, bootstrapping, etc.) provide more stable and robust results and insights, and have therefore become commonplace in ML.
Furthermore, in the case of Quintana et al.\ \cite{quintanaCohortComfortModels2023} and Kim et al.\ \cite{kimPersonalComfortModels2018}, applying Monte Carlo cross-validation to datasets that are both small and highly imbalanced produces results that are opaque and hard to interpret, analyze, or troubleshoot. 
The wide range of performance scores across repeated random splits, in some cases spanning roughly 0.25 to 0.75, is consistent with unstable performance estimates.
This evaluation strategy also obscures the outlier effect we identify in Section~\ref{sec:outliers} (Figs.~\ref{fig:point_clouds_personai_dorn} and \ref{fig:point_clouds_personai_dorn_limited200}, and Figure 5 in Quintana et al.\cite{quintanaCohortComfortModels2023}): a small number of highly imbalanced, trivially-predictable individuals who drive a disproportionate share of the reported gains remain hidden, since in each random split they are always lumped together with a handful of other participants' data in the test set, and their extreme contribution is smoothed out across 100--150 repetitions rather than exposed on a per-person basis.
Leave-one-out cross-validation avoids this problem by producing participant-level results, making it easier to identify the observations driving overall performance and we encourage using it in future personal and cohort comfort modeling.

\subsection{Hypothesis and conclusion}
Quintana et al.~\cite{quintanaCohortComfortModels2023} formulate their central hypothesis as follows: ``our research hypothesis is that the thermal comfort preferences of building occupants can be estimated by analyzing their psychological and behavioral traits and comparing them to other occupants from which subjective feedback was previously obtained.'' 
The original publication does not report a correlation analysis between these traits and thermal preference before proceeding to predictive modelling. 
We addressed this gap in our analysis (Section~\ref{sec:correlation_analysis}) and found only weak linear associations between the measured psychological, demographic, and biological features and thermal preference. 
While the absence of linear associations does not preclude predictive relationships, it provides little initial evidence towards the proposed hypothesis.
The original authors designed ``cold start'' experiments including an ablation analysis, in which participants are assigned to both the correct cohort model and to an incorrect/worst cohort, in order to ``corroborate the usefulness of each cohort approach.'' 
Their original figures (Fig.~4 in \cite{quintanaCohortComfortModels2023}) show that of the six feature sets used for cohorting (sex, surveys, sensitivity, life satisfaction, Big Five personalities, agreeableness), only a single cohorting approach (life satisfaction) shows a marginally higher performance for the correct cohort assignment (0.55 vs.\ 0.53). 
Every other approach performs better when participants are assigned to the wrong cohort; the opposite of what the hypothesis predicts. 
Our own experiments on the DORN and PersonAI datasets show a similar pattern (Section~\ref{sec:classification_performance}, Figs.~\ref{fig:performance_comparison_fig4} and \ref{fig:performance_comparison_fig4_macro}).
In the original publication, the authors discuss the result that supports their hypothesis as follows: ``The Life Satisfaction is the only survey that by itself has an above General-purpose performance. [...] Since both of these cohort approaches are related to satisfaction and optimism, we hypothesize occupants within the resulting cohorts are less prone to preference changes due to small variations, meaning they have a wider acceptability range.''
In contrast, the more common cases in which incorrect cohort assignment outperformed correct assignment receive little discussion. 
The interpretation therefore focuses on evidence consistent with the proposed hypothesis, while the contradictory findings are largely left unconsidered.

Given the small sample size, the small effect sizes, the large number of cohorting strategies, and the absence of formal statistical testing including correction for multiple compoarisons, the study does not establish that this isolated positive finding reflects a genuine effect rather than random variation.
Under these circumstances, a small number of positive results would be expected by chance alone.
The finding should therefore be regarded as exploratory and interpreted cautiously until confirmed by independent replication.
A further result reported by the authors also tempers the interpretation of the Life Satisfaction cohort. 
Table A.2 shows that this approach improved prediction performance for only four participants, while prediction performance decreased for six. 
Thus, even the only cold-start cohorting strategy reported to outperform the general-purpose model on average did not improve prediction performance for most occupants to whom it was applied.
The percentage improvements reported over the General Purpose model for the two cohort approaches with positive average performance are also not examined at the participant level. 
In both cases, the average improvement is driven by two to four participants with very large gains, while the remaining participants show little change or a decrease in performance. 
These large gains occur for participants with highly imbalanced thermal preference distributions (e.g., \texttt{dorn10}, who responded ``prefer cooler'' in 97\% of observations, and \texttt{dorn14}, who responded ``no change'' in 97\% of observations; Figs.~A.1 and A.4 in \cite{quintanaCohortComfortModels2023}), with a similar pattern observed for the SMC dataset (\texttt{smc30} and \texttt{smc46}). Our analysis reproduced the same pattern in both the DORN and PersonAI datasets (Section~\ref{sec:outliers}).

\subsection{Redefinition of metrics}
\label{sec:redefinition_metrics}
Kim et al.\ \cite{kimPersonalComfortModels2018} state that they use ROC-AUC as their evaluation metric, yet throughout the paper the reported performance scores are referred to interchangeably as ``AUC'' and ``accuracy'' due to the lack of familiarity of the term ``AUC'' in the building science community at the time of publication.
While we can see the reasons why authors try to avoid word-repetition in prose text, renaming a well known and well established metric reduces clarity, while introducing the risk to accidentally overlap with another established name of convention ("accuracy").
The resulting terminological imprecisions undermine the reproducibility and comparability of reported results.
However, Kim et al. \cite{kimPersonalComfortModels2018} provides a source and description of the metric and importantly also of the averaging process applied in the multi-class scenario at hand.

\subsection{Information leakage}
\label{sec:data_leakage}
Inspection of the original codebase revealed an information leakage issue arising from the feature selection logic.
Rather than explicitly selecting the intended input variables, the implementation excluded a set of named columns, allowing the dataframe index (\texttt{Unnamed: 0}) to be unintentionally included as a predictor. 
Because both the DORN and SMC datasets span multiple seasons, this index effectively encodes temporal information that should not be available to the model, leading to optimistic performance estimates.
Although we removed this unintended feature in our reproduction, we cannot determine with certainty whether it affected the originally reported results. 
The publicly available code and datasets no longer correspond to a uniquely identifiable version of the original experiments, making it impossible to reconstruct the exact computational environment.
This illustrates a broader challenge as applied disciplines increasingly adopt computational methods: making code and data available is necessary but not sufficient for reproducibility. 
Versioned releases, archived datasets, and commit hashes are also needed to enable independent verification and exact reproduction of reported results [53]. \cite{wilkinsonFAIRGuidingPrinciples2016}.

\subsection{Shared temporal and spatial context between train and test}

\label{sec:temporal_confounding}
Although Quintana et al.\ \cite{quintanaCohortComfortModels2023} perform train-test splits on a participant basis, there is no temporal separation, and only very limited spatial separation, between participants. 
The SMC dataset \cite{kimPersonalComfortModels2018} is collected from participants located in the same office building, and the DORN dataset \cite{tartariniPersonalComfortModels2022} from participants located in the same city. 
Given this spatial proximity, participants assigned to the training and test sets nonetheless share the same weather conditions and seasonal patterns over the same time span, so a model can partly succeed by picking up on shared environmental context rather than on individual-specific comfort preferences. 
A temporal train-test split (e.g., training on an earlier period and testing on a later one) including a deliberate gap in time between the two would better isolate genuine person-level generalization from this shared confounding. We did not correct this issue in our own experiments.

\subsection{Potential overfitting and further information leakage}

\label{sec:hyperparameter_leakage}
Quintana et al.\ \cite{quintanaCohortComfortModels2023} perform hyperparameter tuning on the same data used to evaluate performance, which leaks information from the test set into model selection. 
Tartarini et al.\ \cite{tartariniPersonalComfortModels2022} perform hyperparameter tuning once on the entire dataset and then apply the resulting best parameters across all downstream experiments, which leaks information in a similar way. 
We addressed this issue in our reproduction by removing hyperparameter tuning entirely and relying on the default parameters of scikit-learn's\footnote{scikit-learn 1.6.1} Random Forest implementation.

\section{Conclusion}
We assessed the original methodology of cohort comfort models and conducted two replications using the original dataset and independently collected data, both under a revised experimental setup.
The reported predictive advantages of CCMs were not supported when performance was evaluated using class-imbalance-aware metrics and simple baselines.
More broadly, the limited availability of data remains a fundamental challenge for ML approaches in thermal comfort modeling.
This study highlights the importance of independent replication and rigorous evaluation in assessing the credibility of scientific claims.
Future work should develop guidelines for experimental design and ML analysis in thermal comfort modeling, assess existing methods for personal comfort modeling, and replicate findings from previous personal comfort studies.

\section*{Data Availability}
All data and code used in the experiment is made publicly availabe on Github\footnote{\url{https://github.com/tuw-isab/CCM_Repro_PersonAI/releases/tag/v1.0}} and Zenodo\footnote{\url{todo}}

\section*{Acknowledgements}
The research leading to these results was conducted within the projects  PersonAI (project number FO999901784), funded under the programme ``Technologien und Innovationen für die Klimaneutrale Stadt 2022'', and Wellfit (project number FO999920134), funded under the programme ``Digitale Lösungen für Mensch und Gesellschaft''.  
Both projects were funded by the Austrian Research Promotion Agency (FFG).  
The authors acknowledge TU Wien Bibliothek for financial support  through its Open Access Funding Programme.

\section*{Declaration of generative AI and AI-assisted technologies  in the manuscript preparation process}  
The manuscript structure, scientific arguments, and interpretation  of results were developed entirely by the authors.  
ChatGPT (OpenAI) and Claude (Anthropic) was used for grammatical proofreading and  suggestions to improve sentence structure and readability.  

\appendix
\section{My Appendix}

\printcredits

\bibliographystyle{unsrt}

\bibliography{personaithermalcomfortmodeling}


\end{document}